\pdfoutput=1
\documentclass{gammarobot}
\usepackage[utf8]{inputenc}
\usepackage{xspace,tabularx,array,ragged2e,enumitem,fancyhdr,pifont}
\usetikzlibrary{arrows.meta,calc}
\newcolumntype{L}[1]{>{\RaggedRight\arraybackslash}p{#1}}
\newcolumntype{Y}{>{\RaggedRight\arraybackslash}X}
\setlist[itemize]{noitemsep,leftmargin=*}
\setlist[enumerate]{noitemsep,leftmargin=*}

\fancypagestyle{gammaTeleAI}{%
  \fancyhf{}
  \fancyhead[C]{\small\sffamily Gamma Robotics / TeleAI}
  \fancyfoot[C]{\thepage}
  \renewcommand{\headrulewidth}{\ifnum\value{page}>1 0.3pt\else 0pt\fi}
  
}
\makeatletter
\let\ps@plain\ps@gammaTeleAI
\makeatother

\newcommand{\paperTitle}{DeCoNav-Act: Benchmarking Collective Physical Capability in Multi-Robot Mobile Manipulation}
\newcommand{\bench}{DeCoNav-Act\xspace}
\newcommand{\method}{CoMoMa\xspace}
\newcommand{\ind}{\mathbf{1}}
\newcommand{\res}[1]{#1}
\newcommand{\best}[1]{\textbf{#1}}
\newcommand{\second}[1]{\underline{#1}}
\newcommand{\cmark}{\ding{51}}
\newcommand{\xmark}{\ding{55}}
\newcommand{\eg}{\emph{e.g.}\xspace}
\newcommand{\ie}{\emph{i.e.}\xspace}
\newcommand{\vs}{\emph{vs.}\xspace}
\newcommand{\numHomo}{350\xspace}
\newcommand{\numHetero}{150\xspace}
\newcommand{\numTestHomo}{75\xspace}
\newcommand{\numTestHetero}{25\xspace}
\newcommand{\numTest}{100\xspace}
\newcommand{\numSeeds}{four\xspace}
\newcommand{\numDebugSeeds}{two\xspace}
\newcommand{\numEvalSeeds}{two\xspace}

\hypersetup{pdftitle={\paperTitle},pdfdisplaydoctitle=true}
\title{\paperTitle}
\author[1,2,*]{Yunzi Wu}
\author[3,*]{Sunyao Zhou}
\author[1,4,*]{Xiaotao Zhang}
\author[1,4,*]{Yucheng Deng}
\author[1,5]{Yuehao Huang}
\author[1,6]{Lin Qian}
\author[1,4]{Xuanrui Cui}
\author[1,\dagger]{Xinhai Li}
\author[1,\ddagger]{Chenjia Bai}
\author[2,\ddagger]{Hongqing Chu}
\author[1,\ddagger]{Xuelong Li}

\affiliation[1]{TeleAI}
\affiliation[2]{Tongji University}
\affiliation[3]{Fudan University}
\affiliation[4]{Shanghai Jiao Tong University}
\affiliation[5]{Zhejiang University}
\affiliation[6]{The University of Manchester}

\contribution[*]{Equal contribution}
\contribution[\dagger]{Project leader}
\contribution[\ddagger]{Corresponding authors}
\hypersetup{pdfauthor={\authorlist}}
\date{November 23, 2026}
\metadata[Correspondence to]{\mbox{Yunzi Wu (\email{2611599@tongji.edu.cn})}; \mbox{Chenjia Bai (\email{baicj@chinatelecom.cn})}}
\abstract{%
Robot teams can split workloads or combine complementary bodies, but only if each robot leaves the shared world in a state its teammate can use.
Existing multi-robot benchmarks emphasize task allocation and skill sequencing, rarely testing whether the resulting physical states are usable.
To evaluate \emph{collective physical capability}, we introduce \bench, a benchmark for language-guided multi-robot mobile manipulation in a shared physics simulation.
Its \numHomo homogeneous and \numHetero heterogeneous tasks, each verified by joint reference execution, span two coordination mechanisms: parallel workloads in shared space and physical dependencies in which one robot establishes a condition for a teammate to use.
Beyond team success, \bench measures path efficiency and handoff coverage and success, and adds a diagnostic from verified pre-handoff states.
Evaluating sensor-grounded adaptations of ASPIRE and RAX and our agent \method, we find that with GPT-5.6-luna no method completes a task, at most \res{8.8}\% of dependencies are reached, and even from pre-handoff states no handoff is completed; with GPT-6.1-sol, \method establishes the condition in \res{26.5}\% of trials, but only \res{4.1}\% succeed.
}

\begin{document}
\maketitle
\enlargethispage{\baselineskip}
\suppressfloats[t]
\begin{figure}[t]
    \centering
    \includegraphics[width=0.95\linewidth]{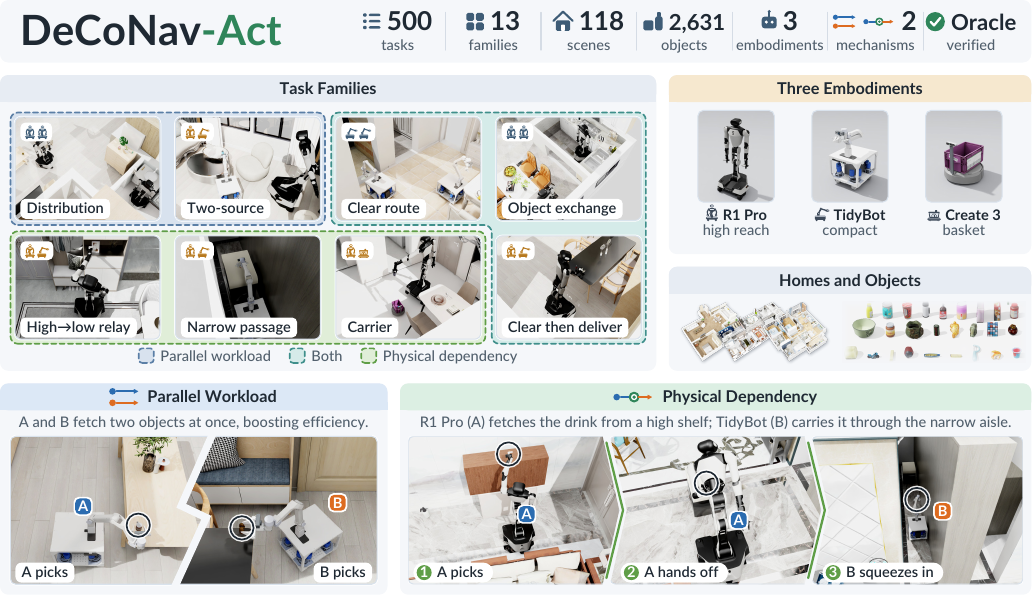}
    \caption{\textbf{Overview of \bench.} \textbf{Top:} key statistics. \textbf{Middle:} eight of the 13 task families grouped by coordination mechanism, the three embodiments, and the VLNVerse homes and $\gamma$-Scape objects. \textbf{Bottom:} two TidyBots pick up objects at separate sources (parallel workload); the R1 Pro (A) leaves a drink within reach of the TidyBot (B), which carries it into a narrow aisle (physical dependency).}
    \label{fig:concept}
\end{figure}

\section{Introduction}
\label{sec:intro}

Household robots increasingly execute long-horizon tasks that couple language, navigation, and manipulation~\citep{behavior,homerobot,embodiedbench}, yet a single robot is bounded by its reach, payload, footprint, and time.
Multiple robots can split a workload or combine different bodies: a tall manipulator retrieves a mug from a high shelf and leaves it where a smaller teammate can grasp it, or loads a carrier that transports several items at once.
These benefits depend on physical execution: the robots share access to counters and passages, and a released object can settle out of the receiver's reach.
Team success therefore depends on \emph{collective physical capability}: the ability of team members to change the shared world in ways their teammates can build on.

Existing benchmarks cover parts of this problem.
Multi-agent mobile manipulation benchmarks emphasize task allocation and sequencing through skill-level interfaces~\citep{partnr,emos,coembod,cocobench,coela,habitat3}.
Physically executed dependencies are evaluated between fixed-base arms on a shared tabletop~\citep{robofactory,roco}, while single-robot mobile manipulation~\citep{behavior,homerobot,maniskillhab} and collaborative navigation~\citep{conavbench,deconav} cover complementary aspects.
These settings leave a gap in evaluating whether a physical state produced by one mobile robot is usable by its teammate in the same evolving scene.

We introduce \bench, a benchmark for language-guided multi-robot mobile manipulation in a shared physics simulation (\cref{fig:concept}).
Tasks are constructed around two coordination mechanisms. In \emph{parallel workload coordination}, robots advance separate goals while sharing sources, destinations, and passages.
In \emph{physical dependency coordination}, one robot establishes a condition for a teammate to use, such as an object released within reach, a payload settled on a carrier, or a cleared route.
Homogeneous teams split workloads and clear shared space for each other; heterogeneous teams add relays and carrier transport that build on their different reach, footprint, and payload.
\bench contains \numHomo homogeneous and \numHetero heterogeneous tasks, each verified by a continuous joint reference execution, with a \numTest-task test split.

Current agents often fail at single-robot navigation or grasping before any cross-robot interaction occurs, so team success alone says little about coordination.
\bench therefore evaluates at two levels.
End-to-end episodes report team success together with per-subgoal path efficiency, the fraction of handoffs reached, and the handoff success rate once attempted.
A handoff-isolated diagnostic starts from verified pre-handoff states, using a reference-producer control to locate failures in establishing usable conditions and completing the consumer's continuation.

We evaluate sensor-grounded, multi-robot adaptations of ASPIRE~\citep{aspire} and RAX~\citep{rax}, with and without the scene prior used by \method, and \method, a reference agent with dependency-aware scheduling, shared spatial memory, and a skill library built during development. All methods use the same sensing, tools, and time budget, and each comparison uses one backbone model.
End to end, no GPT-5.6-luna agent completes a task, and at most \res{8.8}\% of physical dependencies are reached.
With GPT-6.1-sol and shared Oracle guidance, the handoff diagnostic distinguishes agent systems: \method achieves a DSR of \res{33.7}\%, compared with at most \res{14.3}\% for the baselines.
Our contributions are:
\begin{itemize}[leftmargin=*,itemsep=1pt,topsep=2pt]
    \item \bench, a multi-robot mobile manipulation benchmark organized by coordination mechanism, with every task verified through joint physical execution;
    \item a two-level evaluation whose handoff-isolated diagnostic restarts from physically verified states and localizes failures to establishing or using teammate-produced conditions;
    \item an empirical study showing that current agents fail before coordination is tested and that the diagnostic separates coordination designs once low-level inputs are equalized.
\end{itemize}

\section{Related Work}
\label{sec:related}

\begin{table}[tb]
    \caption{\textbf{Comparison with related embodied benchmarks in their main evaluation setting.}
    MR: multiple agents in one scene;
    MM: navigation and object pick-and-place;
    Het.: heterogeneous embodiments;
    Phys.\ dep.: an agent uses an object state that a teammate produced in simulated physics;
    Sensor: no simulator-provided object identities or poses.
    Evidence per cell is in \cref{supp:comparison}.}
    \label{tab:comparison}
    \centering
    \small
    \setlength{\tabcolsep}{4pt}
    \begin{NiceTabular*}{\linewidth}{@{\extracolsep{\fill}}l*{5}{c}@{}}
        \CodeBefore
            \rowcolor[HTML]{F0F8FF}{14}
        \Body
        \toprule
        \textbf{Benchmark}
        & \textbf{MR}
        & \textbf{MM}
        & \textbf{Het.}
        & \textbf{Phys.\ dep.}
        & \textbf{Sensor} \\
        \midrule
        FurnMove~\citep{cordialsync}
        & \cmark & \xmark & \xmark & \xmark & \cmark \\
        TDW-MAT~\citep{coela}
        & \cmark & \cmark & \xmark & \xmark & \cmark \\
        Habitat 3.0~\citep{habitat3}
        & \cmark & \cmark & \cmark & \xmark & \xmark \\
        PARTNR~\citep{partnr}
        & \cmark & \cmark & \cmark & \xmark & \xmark \\
        Habitat-MAS~\citep{emos}
        & \cmark & \cmark & \cmark & \xmark & \xmark \\
        CoEmbod~\citep{coembod}
        & \cmark & \cmark & \xmark & \xmark & \xmark \\
        CoCoBench~\citep{cocobench}
        & \cmark & \cmark & \xmark & \xmark & \xmark \\
        RoCoBench~\citep{roco}
        & \cmark & \xmark & \cmark & \cmark & \xmark \\
        RoboFactory~\citep{robofactory}
        & \cmark & \xmark & \xmark & \cmark & \cmark \\
        BEHAVIOR-1K~\citep{behavior}
        & \xmark & \cmark & \xmark & \xmark & \cmark \\
        CoNavBench~\citep{conavbench}
        & \cmark & \xmark & \xmark & \xmark & \cmark \\
        DeCoNav~\citep{deconav}
        & \cmark & \xmark & \cmark & \xmark & \cmark \\
        \midrule
        \textbf{\bench{} (ours)}
        & \cmark & \cmark & \cmark & \cmark & \cmark \\
        \bottomrule
    \end{NiceTabular*}
\end{table}

\paragraph{Embodied multi-agent collaboration.}
Early collaborative embodied tasks trained agents to jointly find and move furniture with discrete actions~\citep{twobody,cordialsync}.
Later benchmarks study household cooperation with language models, including transport with communication~\citep{coela}, human--robot social rearrangement~\citep{habitat3}, and planning under spatial, temporal, and capability constraints~\citep{partnr}.
Prior work studies task coordination in heterogeneous robot teams and collaborative inference across distributed resources~\citep{emos,coherent,smartllm,llamar,an_aiflow_perspectives,shao_aiflow_edge}; CoEmbod and CoCoBench diagnose allocation, sequencing, mutual exclusion, and inter-agent dependencies~\citep{coembod,cocobench}; and MAGMA-Bench evaluates mission updates and execution failures~\citep{magma}.
At the interaction level, EMOS snaps objects to the gripper with contact physics disabled~\citep{emos}, while PARTNR uses snapping for object grasping and placement~\citep{partnr}.
Recent LLM-based systems transfer objects between real robots~\citep{comuros,robotfleet,closedloopmrm}, but only in a few hand-designed scenarios.
Collaborative manipulation benchmarks evaluate physically executed dependencies among fixed-base arms around a shared tabletop~\citep{roco,lemma,robofactory}.
Collaborative navigation methods and benchmarks study frontier allocation, relay-style handoffs, and dialogue-driven reallocation without manipulation~\citep{conavgpt,conavbench,deconav}.
Among the prior benchmarks in \cref{tab:comparison}, none combines mobile manipulation with physically executed inter-agent dependencies; \bench fills this gap.

\paragraph{Mobile manipulation and embodied evaluation.}
Single-robot benchmarks evaluate rearrangement and long-horizon household activities~\citep{habitat2,homerobot,maniskillhab,behavior}, and EmbodiedBench evaluates multimodal language models as embodied agents~\citep{embodiedbench}.
Beyond binary success, benchmarks report path efficiency~\citep{anderson2018evaluation}, partial goal satisfaction~\citep{behavior}, process-level cooperation scores~\citep{coembod,cocobench}, and skill success from restored intermediate states~\citep{maniunit}.
Our path efficiency applies SPL to individual navigation subgoals, and our handoff-isolated diagnostic adapts ManiUnit's restored-state protocol to cross-robot interactions (\cref{sec:protocol}).
\bench reuses object assets, the BenchForge task format, and the base single-robot Oracle from the concurrently developed $\gamma$-Scape platform~\citep{zhang2026gammascape}.
Our contributions are the team-level construction, joint qualification, heterogeneous families, and two-level evaluation.

\paragraph{Agentic robot control.}
Code as Policies links perception and control with LLM-generated code~\citep{cap}. CaP-X benchmarks coding agents for manipulation~\citep{capx}.
ASPIRE refines robot programs and builds reusable skills through iterative exploration~\citep{aspire}. RAX scales online exploration to generate programs~\citep{rax}.
Navigation policies~\citep{gn0,wnm3d,cogddn,sagenav,sparsevideonav,compass1,navid,uninavid} and vision-language-action models for manipulation~\citep{openvla,pi0} provide complementary single-robot capabilities.
We test how agents combine shared navigation and manipulation tools over long horizons without team-specific demonstrations.

\section{Task Formulation}
\label{sec:task}

We define an episode as
\begin{equation}
 e=(s_0,\mathcal R,\ell,\alpha,\mathcal G,\mathcal D,\mathcal C,\mathcal B),
\end{equation}
where $s_0$ is the initial physical state, $\mathcal R$ the robot team, and $\ell$ a public language instruction.
The instruction names objects, furniture, and rooms, any required handoff surface, and the completion conditions.
The final goals $\mathcal G$ are object-level predicates, $\mathcal D$ are physical dependencies between actions such as grasps, placements, and traversals, $\mathcal C$ are validity constraints such as mutual exclusion at a shared counter, and $\mathcal B$ bounds simulated time, model calls, and tokens.
The reference assignment $\alpha$ records which robot performs each action in the Oracle execution that admits the task (\cref{sec:qualification}); it serves construction and diagnosis and is withheld during end-to-end evaluation.
An episode succeeds if, within $\mathcal B$, its final state satisfies $\mathcal G$ and its trace satisfies $\mathcal D$ and $\mathcal C$, whichever robot performs each action; allocation is thus part of team behavior~\citep{partnr,emos,cocobench}.

\paragraph{Physical dependencies and handoffs.}
A dependency $d=(p,\phi,c)\in\mathcal D$ links a producer action $p$ to a consumer action $c$ through a physical condition $\phi$; under $\alpha$, different robots perform $p$ and $c$.
In a relay, the producer leaves the object on the handoff surface in a stable pose within the receiver's reach, and the receiver grasps it.
In carrier transport, the manipulator settles the payload in the carrier's basket, and the carrier keeps it supported until arrival.
In clearance, the producer clears a destination or passage for the consumer's placement or traversal.
We record establishment $E_d$ ($\phi$ holds when the producer finishes and when the consumer starts) and valid use $U_d$ (the consumer completes $c$ while satisfying its task-specific physical requirements); $d$ is satisfied when both hold, whichever robots act.
A \emph{handoff} is the use of $\phi$ by a robot other than the one that established it; if the producer also performs $c$, $d$ is \emph{self-resolved}, which counts toward team success but not toward handoff success (see \cref{supp:metric_details}) and gives up the team's time advantage (\cref{sec:generation}).
The \emph{pre-handoff state} $s_d^-$ is the state just before the producer's establishing action, \eg, the producer holding the object next to the handoff surface; it anchors the handoff metrics (\cref{sec:metrics}) and the handoff-isolated diagnostic (\cref{sec:diagnostic}).

\paragraph{Coordination mechanisms.}
Under the reference assignment $\alpha$, \emph{parallel workload coordination} arises when robots pursue separate object goals; they may still contend for shared sources, destinations, or passages, which $\mathcal C$ regulates.
Tasks with $\mathcal D\neq\varnothing$ are labeled as involving \emph{physical dependency coordination}.
Clearance tasks keep a parallel workload, since each robot advances its own objects and only the use of the cleared region waits for $\phi$; they exercise both mechanisms.
In relays and carrier transport, the robots instead perform successive stages of the same delivery, linked by $\phi$.
These are the parallel ($\mathcal D=\varnothing$), both, and dependency classes of \cref{fig:construction}c; labels are fixed from the task definition and $\alpha$.
Mechanism and embodiment are labeled separately: homogeneous pairs face dependencies when one robot vacates a destination that its teammate fills, and heterogeneous pairs can share a workload without any handoff.

\section{The \bench Benchmark}
\label{sec:construction}

\bench constructs each task for a designated robot team, verifies it through continuous joint execution, and packages the result with its constraints and reference evidence (\cref{fig:construction}).
Four principles guide its design.
\emph{Physical grounding}: cross-robot physical interactions are executed in continuous physics and judged from robot and object states.
\emph{Mechanism coverage}: tasks span parallel workloads, physical dependencies, and their combination, with homogeneous and heterogeneous teams.
\emph{Verified feasibility}: a task is admitted only after a joint reference execution succeeds.
\emph{Diagnosability}: metrics and controlled conditions locate failures before and during cross-robot interactions.
Because generation and qualification are automated, the same pipeline extends the benchmark to new scenes and robot teams.

\subsection{Simulation, scenes, and robots}
\label{sec:simulation}

All robots and dynamic objects share one physics timeline in Isaac Sim~\citep{isaacsim}, so each robot's actions can change the physical conditions faced by its teammates.
Scenes are furnished multi-room homes from VLNVerse~\citep{vlnverse}, and graspable objects come from $\gamma$-Scape~\citep{zhang2026gammascape}.
Three embodiments have complementary capabilities. The Galaxea R1 Pro is a dual-arm robot with the widest vertical reach.
The TidyBot~\citep{tidybotpp} carries a single arm on a compact base that fits through narrower passages.
The iRobot Create~3 cannot manipulate but carries several items in a basket.
Homogeneous teams pair two TidyBots or two R1 Pros; heterogeneous teams pair the R1 Pro with a TidyBot or a Create~3, and one family uses all three robots.
Specifications are in \cref{supp:robots}.

\subsection{Team-conditioned task generation}
\label{sec:generation}

\begin{figure}[tb]
    \centering
    \includegraphics[width=\textwidth]{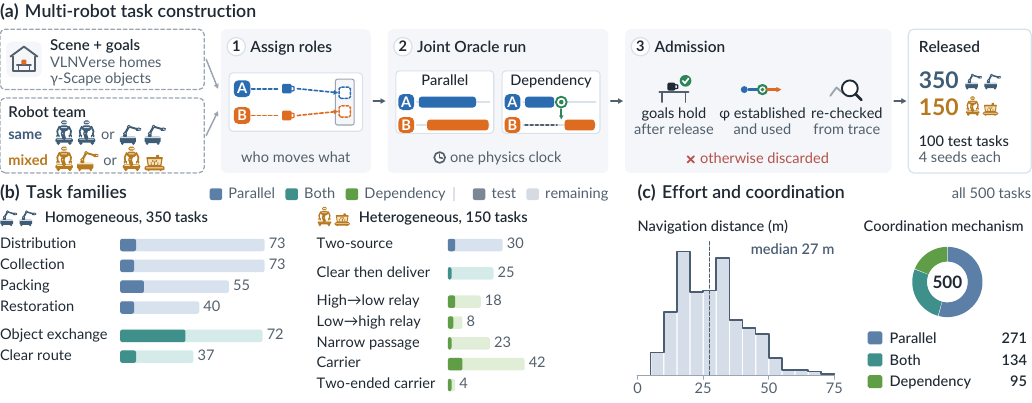}
    \caption{\textbf{Construction and statistics of \bench}. \textbf{(a)} Each candidate binds a scene and its goals to a homogeneous or heterogeneous team and assigns roles. The Oracle executes the whole team on one physics clock; a task is released only if all goals hold, every required physical condition $\phi$ (green node) is established and then used, and an independent audit of the recorded trace confirms these outcomes. \textbf{(b)} Tasks per family, colored by coordination mechanism; darker segments mark the test split. \textbf{(c)} Oracle navigation distance per task, summed over robots, and tasks per coordination mechanism.}
    \label{fig:construction}
\end{figure}

Each candidate binds a scene and its object goals to a robot team, initial states, object assignments, and roles (\cref{fig:construction}a).

\paragraph{Homogeneous tasks.}
Six household families (\cref{fig:construction}b) follow the task objectives of $\gamma$-Scape~\citep{zhang2026gammascape}.
These tasks split workloads between two identical robots; object exchange and clear route also introduce physical dependencies.
The generator searches jointly over object assignments, the second robot's start, and waiting or clearance motions around shared workspaces, ordering actions by the task definition: a robot vacates a destination before its teammate fills it, and a route is cleared before the loaded traversal.
A decomposition is kept only if a parallel-time model predicts that the team finishes before either robot alone.

\paragraph{Heterogeneous tasks.}
Seven families build on differences in reach, footprint, and payload.
Five are dependency-only.
In a \emph{high-to-low relay}, the R1 Pro retrieves an object from elevated storage and leaves it on a handoff surface for the TidyBot; a \emph{low-to-high relay} reverses the roles.
In the \emph{narrow-passage relay}, the R1 Pro transfers an object to the TidyBot, which delivers it through a furniture-lined aisle too narrow for the R1 Pro.
In \emph{carrier transport}, the R1 Pro loads several objects into the Create~3's basket for delivery, and \emph{two-ended loading/unloading} adds a TidyBot that unloads them at the far end.
The other two split workloads: in \emph{clear then deliver}, one robot clears a target surface before its teammate fills it, and \emph{two-source collection} gathers objects at one target.

\paragraph{Instructions.}
Public instructions state the objects, sources, targets, any required handoff surface, and the completion condition, \eg, ``Move the lemonade from the study floor to the westernmost cabinet in the dining room. Use the dining table as the handoff surface.''
They contain no robot-specific steps or coordinates, so agents must ground every referent from their own observations.

\subsection{Joint qualification}
\label{sec:qualification}

\paragraph{Complete-team feasibility.}
Candidates first pass reachability screening for routes, grasps, and placements.
During construction, a privileged Oracle executes the whole team in one scene, planning motions with cuRobo~\citep{curobo} and using the same robot and object models as evaluation.
A task is accepted only if all required actions complete and every final goal holds after release and a stability interval, as confirmed by an independent audit of the recorded states and events rather than by the controller's completion flags (see \cref{supp:qualification}).

\paragraph{Dependency realization.}
For each required dependency, qualification checks both establishment and use.
In a relay, the receiver must acquire and deliver the same object after its partner releases it.
In carrier transport, the payload must settle inside the basket before departure and remain supported until arrival.
In clearance tasks, the vacated region must still be free when the consumer uses it.

\paragraph{Evidence for cooperation.}
The evidence for cooperation depends on the coordination mechanism.
For parallel workloads, a parallel-time model predicts shorter team completion times than for either robot alone; it filters homogeneous candidates (\cref{sec:generation}) and uses measured action durations for heterogeneous collection and clear-then-deliver tasks (see \cref{tab:coop_evidence}).
Reference executions verify feasibility and are not optimized for speed.
For tasks with a physical dependency, qualification verifies that each required dependency is established and used.
For all heterogeneous tasks, Oracle-based qualification verifies both joint feasibility and the necessity of combining complementary robot capabilities within the registered skill set.

\subsection{Statistics and evaluation split}
\label{sec:split}

\Cref{fig:construction}b,c summarizes the \numHomo homogeneous and \numHetero heterogeneous tasks; per-family counts, scenes, and room coverage are in \cref{supp:inventory}.
Of the 500 tasks, 271 exercise only parallel workloads, 95 only physical dependencies, and 134 both; homogeneous tasks fall into the first and third groups (241 and 109), and heterogeneous tasks into all three (30, 95, and 25).
The tasks span 118 scenes and nine room types and contain 2{,}631 object instances.
Per task, the Oracle performs 4.8 manipulation actions (grasps and releases) on average (median 4), and its robots drive a median of 27\,m in total (\cref{fig:construction}c).

\paragraph{Test split.}
The test split contains \numTest comparatively easy tasks that keep long-horizon execution tractable for current agents: \numTestHomo homogeneous tasks with short carrying legs across all six families, and \numTestHetero heterogeneous tasks with at least two per family.
Mobile manipulators start close to their first pickup.
Each test task has \numSeeds seeds that slightly perturb robot starts and object poses, each checked for collision-free and reproducible resets (see \cref{supp:inventory}); \cref{sec:protocol} splits them into withheld and evaluation seeds.
The remaining tasks are released as an additional pool for training and evaluation.

\section{Evaluation Protocol}
\label{sec:protocol}

\bench evaluates teams at two levels: end-to-end episodes from the initial state (\cref{sec:metrics}) and a diagnostic that starts from verified pre-handoff states (\cref{sec:diagnostic}).

\subsection{Execution and information access}
\label{sec:execution}

All robots share one scene and simulation clock. Model requests run asynchronously, while physics continues to advance for all robots and objects, so deliberation time has a physical cost.

The default observation interface is sensor-grounded. Each robot receives its own RGB-D images, proprioception, IMU measurements, camera calibration, and kinematics, together with teammate messages; simulator-provided object poses, scene meshes, traversability maps, reference routes, exact localization, and evaluator predicates are withheld. Some methods additionally receive a \emph{scene prior}: the room layout and static occupancy derived from the scene assets, without the positions of any task object. \method uses it in every end-to-end episode, together with a navigation memory built during development (\cref{sec:methods}). End-to-end episodes provide only the team-level instruction, without prescribed role assignments; roles are specified only in the diagnostic. Recovery is permitted within an episode, but resets are not.
\subsection{Metrics}
\label{sec:metrics}

\paragraph{Team outcomes.}
\emph{Team success} (SR) requires the terminal state to satisfy $\mathcal G$ and the execution trace to satisfy $\mathcal D$ and $\mathcal C$.
\emph{Goal completion} (GC) is the fraction of task checkpoints completed by termination, averaged over episodes.
Checkpoints cover navigation (reaching an object or the loading robot) and manipulation (acquiring, carrying, placing, or loading), and each is credited only after its physical criteria are verified (see \cref{supp:metric_details}). The checkpoint set of a task is fixed and identical for every method; navigation checkpoints already satisfied at reset are excluded.

\paragraph{Path efficiency.}
Each navigation checkpoint defines a leg to its interaction target, which is \emph{reached} when the base of the robot performing the leg arrives near the target (see \cref{supp:metric_details}).
Let $N$ be the number of navigation checkpoints, fixed per task and identical across methods, and $\mathcal A$ the reached subset. With measured planar lengths $L_j$ and reference lengths $L_j^{\rm ref}$,
\begin{equation}
\mathrm{NSR}=\frac{|\mathcal A|}{N},\qquad
\mathrm{PE}=\frac{1}{N}\sum_{j\in\mathcal A}\frac{L_j^{\rm ref}}{\max(L_j^{\rm ref},L_j)}.
\end{equation}
PE is SPL~\citep{anderson2018evaluation} computed per navigation subgoal, \ie, NSR weighted by the efficiency of the reached legs.
The reference length is the shortest collision-free path from the leg's start to the target region, and each leg is attributed to the robot that performs it, so the reference follows the team's own allocation rather than the Oracle's; detours, including those around the teammate, lower PE.

\paragraph{Handoff coverage and success.}
For each dependency $d$ in episode $i$, let $r_{id}=1$ if its pre-handoff conditions are met, whichever robot is the producer, and $x_{id}=1$ if another robot then attempts the consumer action $c$. HCov measures pre-handoff coverage, and HSR measures success once a consumer attempts the handoff:
\begin{equation}
\mathrm{HCov}=\frac{\sum_{i,d} r_{id}}{\sum_i|\mathcal D_i|},\quad
\mathrm{HSR}=\frac{\sum_{i,d} r_{id}\,x_{id}\,\ind[E_{id}\land U_{id}]}{\sum_{i,d} r_{id}\,x_{id}}.
\end{equation}
Stages are credited to the acting robots (see \cref{supp:metric_details}); each metric is undefined when its denominator is zero.

\subsection{Handoff-isolated diagnostic}
\label{sec:diagnostic}

Because most episodes end before any handoff, the diagnostic starts each trial from the pre-handoff state $s_d^-$ of a required dependency, obtained by replaying the joint Oracle reference and physically verified (see \cref{supp:diagnostics}).
The evaluated method receives the instruction and the roles under $\alpha$ and has at most \res{20}\,min of model time.
Diagnostic success rate (DSR) is the fraction of initialized trials in which the producer establishes $\phi$, the consumer uses it, and the consumer completes the prescribed continuation (\eg, placing the relayed object at its target).
Two conditions start from $s_d^-$: in the \emph{full} condition the method controls both robots, and in the \emph{reference-producer} condition the Oracle first performs the producer's establishing action before the method takes over. The establishment rate $E_d$ (full condition) and the valid-use rate $U_d$ thus separate failures in establishing the condition from failures in using it; CKPT is the share of task checkpoints completed in the trial. The diagnostic instantiates \res{107} of the \res{114} test dependencies, omitting the second clearance in seven object-exchange tasks; \res{98} of them yield a verified $s_d^-$ per condition. An \emph{Oracle-guided} setting additionally gives every method Oracle access to object 6D poses, grasp states, and navigation waypoints, to separate coordination from low-level execution.

\subsection{Evaluated methods}
\label{sec:methods}

Following ASPIRE~\citep{aspire}, each test task has \numDebugSeeds withheld seeds, disjoint from its evaluation seeds. \method evolves its skill library and navigation memory on the withheld seeds and freezes both before testing, whereas the baselines have no development phase.
Every method is then tested on one of the \numEvalSeeds evaluation seeds of each test task (\res{100} episodes).

\paragraph{ASPIRE-SG and RAX-SG.}
We adapt ASPIRE~1.0~\citep{aspire,aspirecode} and RAX~\citep{rax,raxcode} to multi-robot teams and to the sensor-grounded interface, replacing privileged inputs of their original implementations, such as exact odometry, traversability maps, and simulator collision worlds, with the shared tools.
Each robot runs its own agent with the upstream executor, and the agents divide the work through the same team board and message tools as \method; no exploration is run before testing (see \cref{supp:adapters}).
We evaluate each baseline with and without the scene prior (denoted +prior in \cref{tab:main}).

\paragraph{\method.}
\method builds on the program-as-policy design of ASPIRE and RAX: its runner follows ASPIRE's tool registry and program executor, and robot actions are composed from registered tools and revised from execution feedback.
Both target single-robot control; \method adds three components for team execution.
(i)~\emph{Collaboration contract.} A shared task graph records the assigned robot and prerequisites of every subtask; a subtask becomes executable only after its prerequisites are verified, and operations on shared objects, workspaces, and handoff surfaces acquire leases.
(ii)~\emph{Shared spatial memory.} Measured destinations, maps, and routes are shared with their provenance and kept separate from the task graph, so a teammate's progress updates only the affected bindings. A navigation memory of measured routes, compiled from development episodes, proposes subgoals to the current-map planner.
(iii)~\emph{Development-time skill library.} A skill library is refined offline from development failures (see \cref{supp:comoma}).
The +prior baselines share \method's tools and scene prior, so their gap to \method reflects components (i)--(iii) together.

\paragraph{Shared settings.}
All methods use \res{GPT-5.6-luna} with \res{medium} reasoning effort, the same tools, and a per-episode budget of \res{20}\,min of model time and \res{500}\,s of simulated time (see \cref{supp:budgets}).
\res{We additionally run \method end to end, and all methods in the diagnostic, with GPT-6.1-sol (\cref{tab:diag}).}

\section{Experiments}
\label{sec:experiments}

We address four questions. \textbf{Q1}: How far do current agents get end to end? \textbf{Q2}: Where does execution stop? \textbf{Q3}: From verified pre-handoff states, which handoff stage fails, and does the diagnostic separate coordination designs? \textbf{Q4}: How much do the scene prior, \method's components, and the backbone contribute? \Cref{tab:main,tab:diag} report the end-to-end and handoff-isolated results.

\begin{table}[!b]
    \caption{\textbf{End-to-end results (\%).}
    One episode per test task and method (\res{100} per method, covering \res{878} checkpoints and \res{114} dependencies); metrics follow \cref{sec:metrics}.
    The baselines adapt ASPIRE~\citep{aspire} and RAX~\citep{rax}. All rows use GPT-5.6-luna except the last.
    HSR is shown as successful/attempted handoffs (--: none attempted).
    \best{Bold} / \second{underline}: best / second best among GPT-5.6-luna methods.}
    \label{tab:main}
    \centering
    \small
    \setlength{\tabcolsep}{3pt}
    \begin{NiceTabular*}{\textwidth}{@{\extracolsep{\fill}}l*{12}{c}@{}}
        \CodeBefore
            \rowcolor[HTML]{F0F8FF}{7,8}
        \Body
        \toprule
        \Block{2-1}{\textbf{Method}}
        & \multicolumn{6}{c}{\textbf{All tasks (100)}}
        & \multicolumn{3}{c}{\textbf{Homogeneous (75)}}
        & \multicolumn{3}{c}{\textbf{Heterogeneous (25)}} \\
        \cmidrule(lr){2-7}
        \cmidrule(lr){8-10}
        \cmidrule(lr){11-13}
        & \textbf{SR}$\uparrow$ & \textbf{GC}$\uparrow$ & \textbf{NSR}$\uparrow$ & \textbf{PE}$\uparrow$ & \textbf{HCov}$\uparrow$ & \textbf{HSR}$\uparrow$
        & \textbf{GC}$\uparrow$ & \textbf{NSR}$\uparrow$ & \textbf{HCov}$\uparrow$
        & \textbf{GC}$\uparrow$ & \textbf{NSR}$\uparrow$ & \textbf{HCov}$\uparrow$ \\
        \midrule
        ASPIRE-SG
        & 0.0 & 3.2 & 11.0 & 9.9 & 0.9 & --
        & 4.2 & 16.3 & 1.2
        & 0.0 & 0.0 & 0.0 \\
        RAX-SG
        & 0.0 & 2.9 & 10.3 & 8.9 & 0.0 & --
        & 3.8 & 15.4 & 0.0
        & 0.0 & 0.0 & 0.0 \\
        ASPIRE-SG + prior
        & 0.0 & 9.6 & \second{41.9} & \second{35.5} & 1.8 & --
        & 10.1 & 51.0 & 2.5
        & \second{8.0} & \second{23.5} & 0.0 \\
        RAX-SG + prior
        & 0.0 & \second{9.9} & 40.0 & 33.3 & \second{4.4} & 0/1
        & \second{11.8} & \second{53.8} & \second{6.2}
        & 4.3 & 11.8 & 0.0 \\
        \textbf{\method}
        & 0.0 & \best{13.2} & \best{54.2} & \best{48.6} & \best{8.8} & \best{1/5}
        & \best{13.5} & \best{59.6} & \best{12.3}
        & \best{12.3} & \best{43.1} & 0.0 \\
        \midrule
        \textbf{\method} (GPT-6.1-sol)
        & 1.0 & 21.5 & 60.6 & 51.8 & 23.7 & 4/11
        & 20.5 & 67.3 & 19.8
        & 24.6 & 47.1 & 33.3 \\
        \bottomrule
    \end{NiceTabular*}
\end{table}

\subsection{End-to-end evaluation}
\label{sec:e2e}

\paragraph{Team success remains out of reach (Q1).}
With GPT-5.6-luna, no method completes a single task, and even the strongest configuration, \method, reaches only \res{13.2}\% GC (\cref{tab:main}). A stronger backbone helps without changing the picture: with GPT-6.1-sol, \method completes one of \res{100} tasks, a carrier task in which the R1~Pro loads both objects into the Create~3's basket. \bench thus leaves ample headroom, and its graded metrics separate the methods (\method over the strongest baseline: \res{+3.3} GC points, 95\% CI [\res{0.5}, \res{6.2}]).

\paragraph{Execution stops before most handoffs (Q2).}
Agents reach navigation targets far more often than they complete manipulation: with the scene prior, about half of the navigation checkpoints are reached, but fewer than \res{4}\% of the manipulation checkpoints are completed, so GC mainly reflects navigation. Few dependencies are therefore reached, and consumers attempt only \res{6} handoffs in \res{500} GPT-5.6-luna episodes. PE stays within \res{83}--\res{90}\% of NSR, so navigation fails by not arriving rather than by detours. End-to-end HSR thus rests on few attempts, which motivates the handoff-isolated diagnostic.

\paragraph{Physical dependencies are harder than parallel workloads.}
At matched embodiment, homogeneous tasks with physical dependencies roughly halve the GC of the +prior baselines relative to parallel-only tasks (\eg, from \res{16.2}\% to \res{8.1}\% for RAX-SG + prior), whereas \method's GC is unaffected, consistent with its dependency-aware task graph. Among heterogeneous families, relays are the hardest: no GPT-5.6-luna method credits any checkpoint on them, and only GPT-6.1-sol makes progress, while carrier transport is the most tractable. Per-family results are in \cref{tab:per_family}.

\paragraph{The scene prior matters most; the backbone helps manipulation (Q4).}
The scene prior, the largest single factor, roughly quadruples the baselines' NSR and triples their GC. Given the same prior and tools, \method's further gain concentrates on heterogeneous tasks and dependencies, where its HCov doubles that of the best baseline; this gap is attributable to \method's team-level design as a whole, namely its coordination layer, navigation memory, and skill library (\cref{sec:methods}). Switching to GPT-6.1-sol raises manipulation completion from \res{3.7}\% to \res{12.2}\% and HCov to \res{23.7}\%, and consumers then complete \res{4} of \res{11} attempted handoffs.

\subsection{Handoff-isolated diagnostic (Q3)}
\label{sec:diag_results}
\begin{table}[tb]
    \caption{\textbf{Handoff-isolated diagnostic (\%).}
    \res{98} trials per method and condition; conditions and metrics follow \cref{sec:diagnostic}.
    (a) GPT-5.6-luna without the scene prior;
    (b) GPT-5.6-luna and (c) GPT-6.1-sol, where only \method uses the scene prior and navigation memory;
    (b) and (c) share one software version;
    (d) as (c), Oracle-guided.}
    \label{tab:diag}
    \centering
    \small
    \setlength{\tabcolsep}{2.5pt}
    \begin{NiceTabular*}{\columnwidth}{@{\extracolsep{\fill}}cl*{7}{c}@{}}
        \CodeBefore
            \rowcolor[HTML]{F0F8FF}{5,8,11,14}
        \Body
        \toprule
        & \Block{2-1}{\textbf{Method}}
        & \multicolumn{4}{c}{\textbf{Full}}
        & \multicolumn{3}{c}{\textbf{Ref.\ prod.}} \\
        \cmidrule(lr){3-6}
        \cmidrule(lr){7-9}
        & & \textbf{CKPT} & {\boldmath$E_d$} & {\boldmath$U_d$} & \textbf{DSR}
        & \textbf{CKPT} & {\boldmath$U_d$} & \textbf{DSR} \\
        \midrule
        \Block{3-1}{(a)}
        & ASPIRE-SG & 3.2 & \best{4.1} & 0.0 & 0.0 & 3.4 & 0.0 & 0.0 \\
        & RAX-SG    & 3.9 & 2.0 & 0.0 & 0.0 & \best{3.7} & \best{1.0} & 0.0 \\
        & \textbf{\method} & \best{5.8} & 0.0 & 0.0 & 0.0 & \best{3.7} & \best{1.0} & 0.0 \\
        \midrule
        \Block{3-1}{(b)}
        & ASPIRE-SG & 3.3 & 0.0 & 0.0 & 0.0 & 2.7 & 0.0 & 0.0 \\
        & RAX-SG    & 4.1 & 1.0 & 0.0 & 0.0 & 3.5 & 0.0 & 0.0 \\
        & \textbf{\method} & \best{5.4} & \best{4.1} & 0.0 & 0.0 & \best{4.7} & \best{4.1} & 0.0 \\
        \midrule
        \Block{3-1}{(c)}
        & ASPIRE-SG & 5.8 & 4.1 & 0.0 & 0.0 & 5.9 & 4.1 & 1.0 \\
        & RAX-SG    & 7.6 & 7.1 & 1.0 & 0.0 & 8.1 & 7.1 & 2.0 \\
        & \textbf{\method} & \best{17.3} & \best{26.5} & \best{9.2} & \best{4.1} & \best{11.5} & \best{22.4} & \best{7.1} \\
        \midrule
        \Block{3-1}{(d)}
        & ASPIRE-SG & 22.9 & 33.7 & 23.5 & 13.3 & 14.0 & 28.6 & 18.4 \\
        & RAX-SG    & 24.8 & 39.8 & 24.5 & 14.3 & 21.2 & 39.8 & 29.6 \\
        & \textbf{\method} & \best{33.4} & \best{56.1} & \best{39.8} & \best{33.7} & \best{28.9} & \best{64.3} & \best{49.0} \\
        \bottomrule
    \end{NiceTabular*}
\end{table}

\paragraph{Without guidance, few handoffs succeed.}
With GPT-5.6-luna, no method completes a handoff in either condition (settings (a) and (b) in \cref{tab:diag}): producers rarely establish the condition, and consumers rarely use one that the Oracle establishes, with or without the scene prior and navigation memory. With GPT-6.1-sol (setting (c)), \method's producer establishes the condition in about a quarter of full-condition trials, yet only \res{4.1}\% complete the handoff, and an Oracle-established condition raises this only to \res{7.1}\%; the baselines gain much less. Establishment thus remains the largest loss, but the stronger backbone exposes a second one: \res{17} of the \res{26} conditions that \method establishes are not validly used.

\paragraph{With low-level inputs equalized, the diagnostic separates coordination designs.}
To test whether the diagnostic measures coordination rather than low-level execution, setting (d) repeats (c) in the Oracle-guided setting (\cref{sec:diagnostic}), which gives every method the same object poses, grasp states, and navigation waypoints. All methods improve, and \method's margin widens: its full-condition DSR reaches \res{33.7}\%, against \res{14.3}\% for RAX-SG and \res{13.3}\% for ASPIRE-SG, and \res{49.0}\% against at most \res{29.6}\% with the Oracle producer (paired McNemar test, $p<0.05$ for every comparison). The stage-wise rates localize the gap: \method establishes the condition more often (\res{56.1}\% \vs at most \res{39.8}\%) and completes the continuation in \res{85}\% of the trials in which it uses the condition, against \res{57}--\res{58}\% for the baselines, whereas using an established condition is similarly hard for all methods (\res{62}--\res{71}\%). Even \method fails to establish the condition in \res{43} of \res{98} trials, so both stages remain open.

\subsection{Real-world case study}
\label{sec:real_world}
\begin{figure}[tb]
    \centering
    \includegraphics[width=\textwidth]{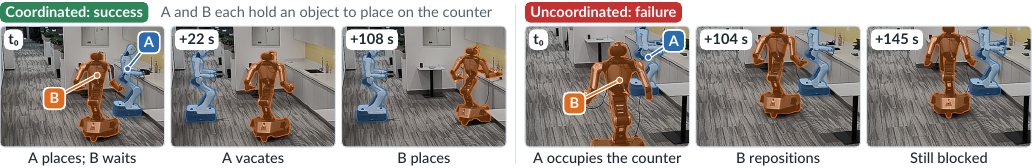}
    \caption{\textbf{Real-world shared-counter coordination.} An Astribot A and an R1~Pro B, highlighted in blue and orange, each bring an object to the same counter. \textbf{Left:} with coordination, A places its object and vacates the counter, so B can place the second object. \textbf{Right:} without coordination, A keeps the counter occupied and B stays blocked. $t_0$ is the initial time, and $+n$\,s is the time elapsed since $t_0$.}
    \label{fig:realworld}
\end{figure}

We test whether these constraints arise on hardware with a shared-counter task for an Astribot and an R1~Pro (\cref{fig:realworld}), in which one robot's presence blocks the other's access to the counter. With coordination, the Astribot announces its placement and vacates the counter, after which the R1~Pro places its object; without these messages, the Astribot keeps occupying the counter, and the R1~Pro repeatedly repositions without placing. Neither robot fails to grasp or navigate, so releasing a shared resource is itself a physical precondition for the teammate's progress, which \bench encodes as a validity constraint.

\subsection{Failure analysis}
\label{sec:failures}

\paragraph{End to end, single-robot stages fail first.}
Nearly every GPT-5.6-luna failure occurs before a cross-robot interaction. In the tool logs of \method and the baselines without the prior, about half of \method's navigation requests arrive, against roughly one in eight for the baselines, and only \res{11} of \res{451} point-grasp requests yield a held object. Failed grasps stem mostly from motion execution, such as inverse kinematics or rejected arm paths. The diagnostic by dependency type is in \cref{tab:diag_type}.

\paragraph{Takeaways.}
End-to-end success is bounded by single-robot navigation and grasping. With low-level inputs equalized, the diagnostic separates coordination designs and localizes their failures to establishing and continuing from teammate-produced conditions; most handoffs remain unsolved even for the strongest agent, which calls for agents that reason explicitly about the states they leave for teammates.

\section{Conclusion}
\label{sec:conclusion}

We introduced \bench, a benchmark for language-guided multi-robot mobile manipulation in a shared physical world, with tasks organized by coordination mechanism across homogeneous and heterogeneous teams and admitted through continuous joint execution. Its two-level evaluation shows that current agents rarely reach a handoff end to end and that its diagnostic separates coordination designs once low-level inputs are equalized, while establishing and using teammate-produced conditions remain largely unsolved. We hope \bench helps measure progress toward collective physical capability, and we plan to extend it with longer horizons, larger teams, and task allocation.

\clearpage
\pdfbookmark[1]{\refname}{references}
\bibliographystyle{plainnat}
\bibliography{references}

\clearpage
\beginappendix
\section{Task Families, Robots, and Qualification}
\label{supp:tasks}
\subsection{Inventory and spatial coverage}
\label{supp:inventory}
The homogeneous inventory contains 350 parent configurations, selected from 427 candidates that passed shared-world Oracle execution, completion of all required actions, and post-release stability. The selection balances family sizes and preserves coverage; it keeps all 116 candidate scenes and 513 of 574 object assets. The 75 homogeneous test tasks were fixed before this selection, ranked by the longer of the two robots' first carrying legs and then by their sum. Among them, 34 tasks contain more than one transport cycle and 32 require an object released by one robot to be regrasped by the other. The heterogeneous inventory contains 150 tasks across all seven families (including 23 furnished-aisle narrow-passage tasks and per-family quotas that cap carrier transport at 42 tasks), selected from semantically distinct candidates that passed continuous joint execution and the independent trace audit. The heterogeneous tasks span 19 scenes and 36 rooms of five types and use 61 distinct object assets; 41\% of them span more than one room. Seed and route variants sharing the same scene, objects, source, handoff, and target furniture, and instruction are deduplicated and never counted as separate tasks, nor are the \numSeeds initial-state seeds per test task.
\Cref{tab:families} lists the per-family counts shown in \cref{fig:construction}(b) together with scene counts.

\begin{table}[!b]
\centering\small
\caption{Benchmark and test-split statistics (parent configurations; seeds not counted). H: homogeneous pair; X: heterogeneous team. The bottom rows give instance-level mechanism labels. Room coverage is in \cref{tab:room_coverage}.}
\label{tab:families}
\setlength{\tabcolsep}{3pt}
\begin{tabular}{@{}lcrrc@{}}
\toprule
Task family & Team & Full & Test & Scenes (F/T) \\
\midrule
\multicolumn{5}{@{}l}{\textit{Workload sharing}} \\
Multi-destination distrib. & H & 73 & 8 & 62 / 6 \\
Multi-source collection & H & 73 & 7 & 52 / 7 \\
Container packing & H & 55 & 12 & 46 / 12 \\
Environment restoration & H & 40 & 7 & 34 / 6 \\
Two-source collection & X & 30 & 4 & 5 / 4 \\
\multicolumn{5}{@{}l}{\textit{Clearance and exchange}} \\
Object exchange & H & 72 & 33 & 68 / 29 \\
Clear route and deliver & H & 37 & 8 & 29 / 6 \\
Clear then deliver & X & 25 & 2 & 1 / 1 \\
\multicolumn{5}{@{}l}{\textit{Object relay}} \\
High-to-low relay & X & 18 & 4 & 2 / 2 \\
Low-to-high relay & X & 8 & 3 & 2 / 2 \\
Narrow-passage relay$^\dagger$ & X & 23 & 2 & 5 / 2 \\
\multicolumn{5}{@{}l}{\textit{Carrier transport}} \\
Multi-object carrier & X & 42 & 8 & 14 / 8 \\
Two-ended load/unload & X & 4 & 2 & 1 / 1 \\
\midrule
Homogeneous total & H & \textbf{350} & \textbf{75} & 116 / 50 \\
Heterogeneous total & X & \textbf{150} & \textbf{25} & 19 / 14 \\
\midrule
\multirow{2}{*}{Parallel only ($\mathcal D=\varnothing$)} & H & 241 & 34 & -- \\
 & X & 30 & 4 & -- \\
\multirow{2}{*}{With dependency ($\mathcal D\neq\varnothing$)} & H & 109 & 41 & -- \\
 & X & 120 & 21 & -- \\
\bottomrule
\end{tabular}
\\[2pt]
{\footnotesize $^\dagger$Aisles lined with original-size household furniture along an existing wall.}
\end{table}

\begin{table}[tb]
\centering\small
\caption{Interaction-room coverage. Homogeneous cells report full / test counts; heterogeneous rows report test counts, except for the last row, which covers the full heterogeneous inventory. A task contributes at most once per room type, and scene totals are unions rather than sums over families.}
\label{tab:room_coverage}
\setlength{\tabcolsep}{3pt}
\begin{tabular*}{\textwidth}{@{\extracolsep{\fill}}lrrrrrrrrrr@{}}
\toprule
\textbf{Task family} & \textbf{Scenes} & \textbf{Liv.} & \textbf{Kit.} & \textbf{Bed.} & \textbf{Bath} & \textbf{Study} & \textbf{Din.} & \textbf{Bal.} & \textbf{Lau.} & \textbf{Stor.} \\
\midrule
Multi-destination distribution & 62/6 & 69/6 & 19/1 & 16/2 & 18/2 & 13/2 & 5/2 & 5/1 & 1/0 & 0/0 \\
Multi-source collection & 52/7 & 70/7 & 30/4 & 21/0 & 6/1 & 10/0 & 4/2 & 4/0 & 1/0 & 0/0 \\
Container packing & 46/12 & 46/10 & 17/6 & 20/1 & 15/1 & 7/2 & 2/1 & 2/2 & 0/0 & 1/1 \\
Environment restoration & 34/6 & 35/7 & 16/4 & 7/0 & 4/1 & 10/0 & 4/2 & 3/0 & 1/0 & 0/0 \\
Object exchange & 68/29 & 48/24 & 37/17 & 14/4 & 18/9 & 14/7 & 8/4 & 5/1 & 0/0 & 0/0 \\
Clear route and deliver & 29/6 & 27/7 & 10/4 & 13/0 & 3/1 & 18/4 & 1/0 & 0/0 & 0/0 & 0/0 \\
\midrule
\textbf{Homogeneous total} & 116/50 & 295/61 & 129/36 & 91/7 & 64/15 & 72/15 & 24/11 & 19/4 & 3/0 & 1/1 \\
\midrule
\multicolumn{11}{@{}l}{\textit{Heterogeneous test split}} \\
Two-source collection & 4 & 3 & 0 & 1 & 0 & 0 & 0 & 0 & 0 & 0 \\
Clear then deliver & 1 & 2 & 0 & 0 & 0 & 0 & 0 & 0 & 0 & 0 \\
High-to-low relay & 2 & 4 & 0 & 0 & 0 & 0 & 1 & 0 & 0 & 0 \\
Low-to-high relay & 2 & 2 & 0 & 0 & 0 & 1 & 1 & 0 & 0 & 0 \\
Narrow-passage relay & 2 & 2 & 0 & 0 & 0 & 0 & 0 & 0 & 0 & 0 \\
Multi-object carrier transport & 8 & 7 & 0 & 7 & 1 & 1 & 0 & 0 & 0 & 0 \\
Two-ended loading/unloading & 1 & 2 & 0 & 2 & 0 & 0 & 0 & 0 & 0 & 0 \\
\midrule
\textbf{Heterogeneous test total} & 14 & 22 & 0 & 10 & 1 & 2 & 2 & 0 & 0 & 0 \\
\textbf{Heterogeneous full inventory} & 19 & 130 & 0 & 55 & 2 & 11 & 13 & 0 & 0 & 0 \\
\bottomrule
\end{tabular*}
\vspace{3pt}
\begin{minipage}{\textwidth}\footnotesize
Liv.: living room; Kit.: kitchen; Bed.: bedroom; Bath: bathroom; Din.: dining room; Bal.: balcony; Lau.: laundry room; Stor.: storage. Room labels are normalized by case and whitespace. Zeros indicate that no interaction workspace of that type is used, not that the room is absent from the scene.
\end{minipage}
\end{table}

\paragraph{Spatial counting unit.}
Scene coverage counts unique source scenes. Room coverage counts the grounded interaction workspaces of each task, namely object sources, handoff surfaces, and targets (for heterogeneous tasks, object sources and targets only); rooms that a robot only passes through are not counted. The homogeneous benchmark covers nine room types and its test split eight (all except laundry rooms), with the 100-task test split drawn as a stratified subset of the 500-task benchmark.

\paragraph{Heterogeneous test split.}
The 25 heterogeneous test tasks cover all seven families, 14 scenes, 25 rooms, and 38 object assets; 12 tasks span more than one room. Each family has at least two test tasks and carrier transport is limited to eight; within a family, selection favors uncovered scenes and short carrying routes. Their public instructions contain 29--56 words (median 41).

\paragraph{Mechanism classes and precedence.}
Mechanism classes follow the required dependencies $\mathcal D$ of each task, not the order in which actions happen. Parallel tasks may still require ordered access to a shared support, and relays can permit parallel preparation. Among the 75 homogeneous test tasks, 34 are parallel, 26 of them without any cross-robot precedence and eight with completion-order constraints at a shared support, and 41 combine a parallel workload with an exchange or clearance dependency.

\paragraph{Test split and seeds.}
The 75 homogeneous test tasks span 50 scenes and eight room types; the median of the longer of the two robots' first carrying legs is 5.16\,m, and 59 tasks use two TidyBots and 16 two R1 Pros. Among the 25 heterogeneous test tasks, 15 pair the R1 Pro with a TidyBot, eight with a Create~3, and two use all three robots; their mean carrying route is 10.2\,m. Initial approaches to the first pickup average 1.54\,m (homogeneous) and 1.82\,m (heterogeneous). Each seed fixes one instance of the task, including the robots' initial poses and the object placements; seeds slightly perturb these initial states and are admitted only if the starts are collision-free and a fresh reset reproduces the configured state. Development and evaluation seeds are disjoint.

\subsection{Robots}
\label{supp:robots}

\begin{table}[tb]
\centering\small
\caption{Robot embodiments. All robots run in Isaac Sim~5.1 on a shared 120\,Hz physics timeline.}
\label{tab:robots}
\begin{tabularx}{\textwidth}{@{}L{0.16\textwidth}L{0.22\textwidth}L{0.24\textwidth}Y@{}}
\toprule
Robot & Base & Manipulation & Capability used in task design \\
\midrule
Galaxea R1 Pro & Omnidirectional wheeled base; 4-DoF torso & Two 7-DoF arms with parallel grippers & Highest vertical reach; retrieves from and places on elevated storage; loads the carrier \\
TidyBot + ARX5 & Compact planar mobile base & One 6-DoF ARX5 arm with parallel gripper & Fits through furniture-lined narrow aisles; low-source and receiver roles \\
iRobot Create~3 & Differential drive, 341.9\,mm footprint length & None & Transports several items in a basket \\
\bottomrule
\end{tabularx}
\end{table}

The three embodiments differ in reach, footprint, and payload (\cref{tab:robots}). Homogeneous tasks use two TidyBots (59 test tasks) or two R1 Pros (16 test tasks). Heterogeneous tasks pair the R1 Pro with a TidyBot (relays, collection, clearance) or with a Create~3 (carrier transport); the two-ended family uses all three robots. The Oracle adapters for the TidyBot--ARX5 and the Create~3 are part of this work.

\subsection{Public task and private evaluator specification}
Each task has a stable identifier, a scene and asset revision, an embodiment configuration, initial-state seeds, a public instruction, a role assignment, final-goal definitions, required interactions, and budgets. The public task conveys goals; the role assignment serves construction, auditing, and the handoff-isolated diagnostic and is withheld in end-to-end episodes. Neither contains object-instance handles, simulator paths, exact coordinates, an executable reference plan, or evaluator state. Agents may create opaque handles from their own detections.

The role assignment specifies an object or action responsibility, not a trajectory. In the diagnostic, a relay receiver is told which object it must take from the handoff surface, but not the exact release pose or when that pose becomes usable. Final goals are object-level outcomes: a mug and a spoon delivered to a desk are two goals, not eight successes for finding, grasping, carrying, and placing. A handoff is required only when the task contract names it, for example through a designated handoff surface.

\subsection{Representative tasks}
\newcommand{\exframe}[3]{\parbox[t]{0.243\textwidth}{\vspace{0pt}\includegraphics[width=\linewidth]{figures/supp_examples/#1}\\[1pt]\centering\textbf{#2}~#3\par}}
\newcommand{\extask}[3]{\multicolumn{4}{@{}p{\textwidth}@{}}{\textbf{#1} (#2).\enspace\textit{``#3''}}\\\noalign{\vskip 2pt}}

\begin{figure}[tb]
\centering\scriptsize
\setlength{\tabcolsep}{1.5pt}
\begin{tabular}{@{}llll@{}}
\extask{Multi-destination distribution}{two TidyBots; parallel workload with a shared source}{Move the mug from the table in the balcony to the dining table in the living room, and move the candle from the table in the balcony to the table in the living room.}
\exframe{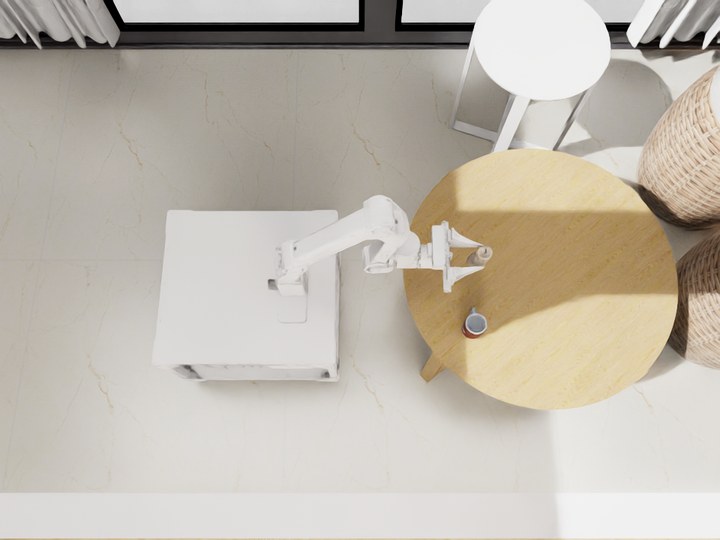}{1}{One TidyBot picks an item at the shared table}
& \exframe{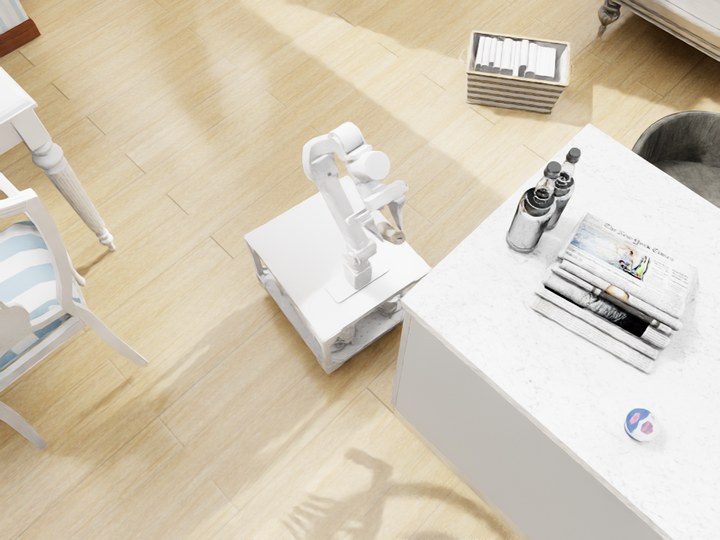}{2}{It delivers the item in the living room}
& \exframe{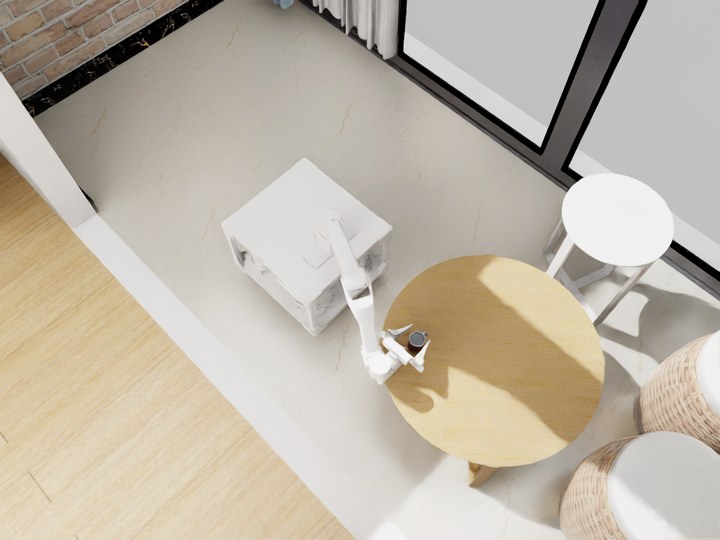}{3}{The other TidyBot picks the second item}
& \exframe{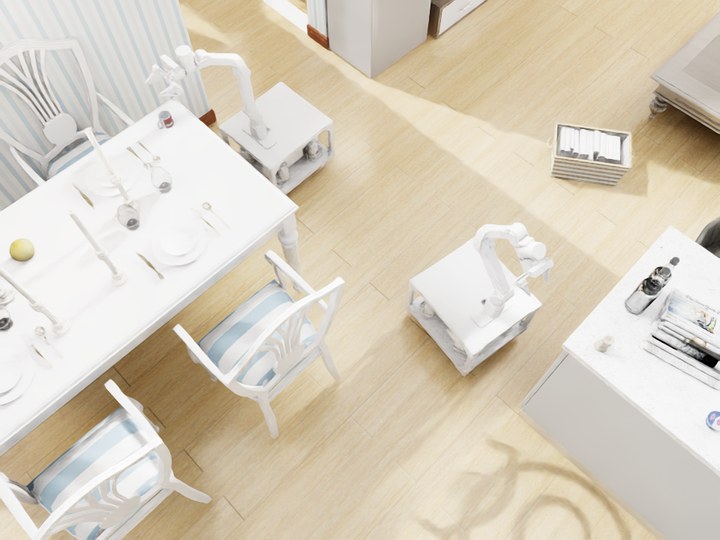}{4}{Both items are delivered}\\\noalign{\vskip 8pt}
\extask{Two-source collection}{R1 Pro and TidyBot; parallel workload}{Put the peanut butter jar and pitcher from the westernmost bedroom cabinet and the can and water bottle from the table closest to the bed onto the cabinet nearest the television in the bedroom. Leave all four items upright, released, and stable.}
\exframe{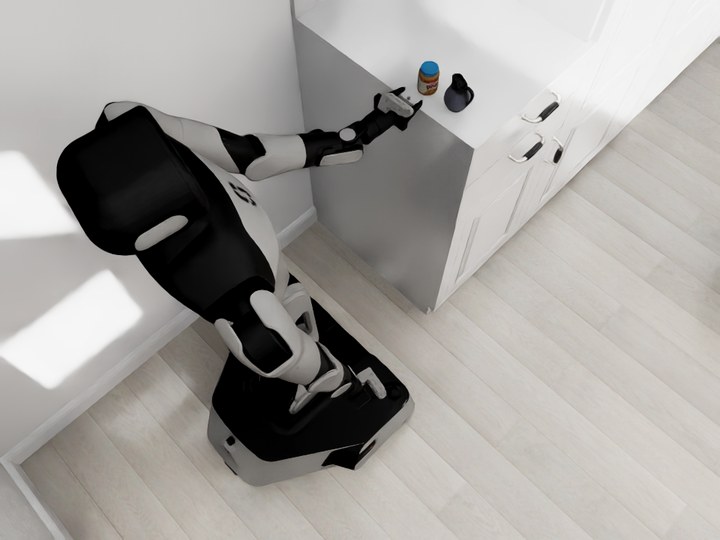}{1}{R1 Pro picks from the tall cabinet}
& \exframe{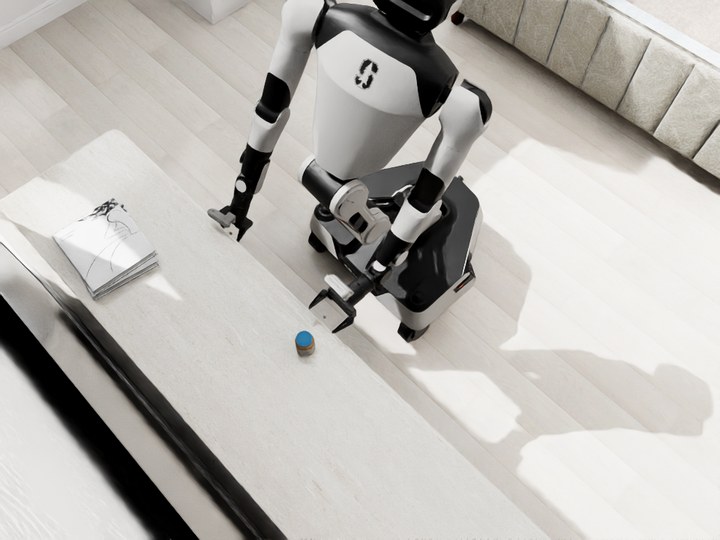}{2}{R1 Pro places on the TV cabinet}
& \exframe{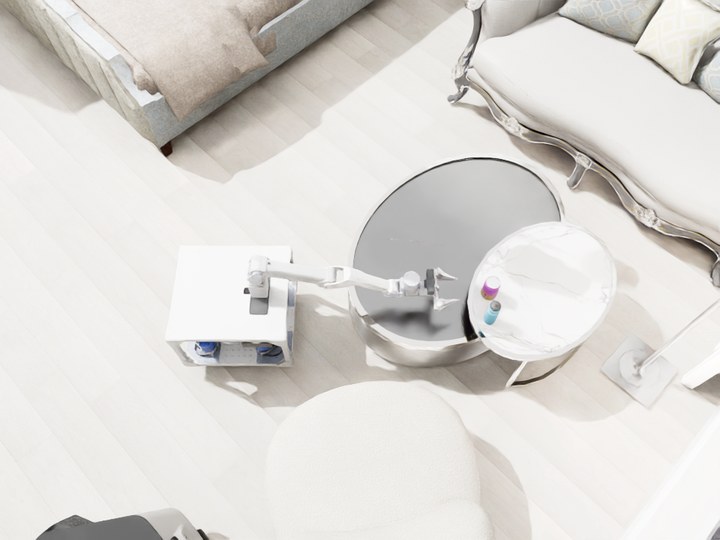}{3}{TidyBot picks from the low table}
& \exframe{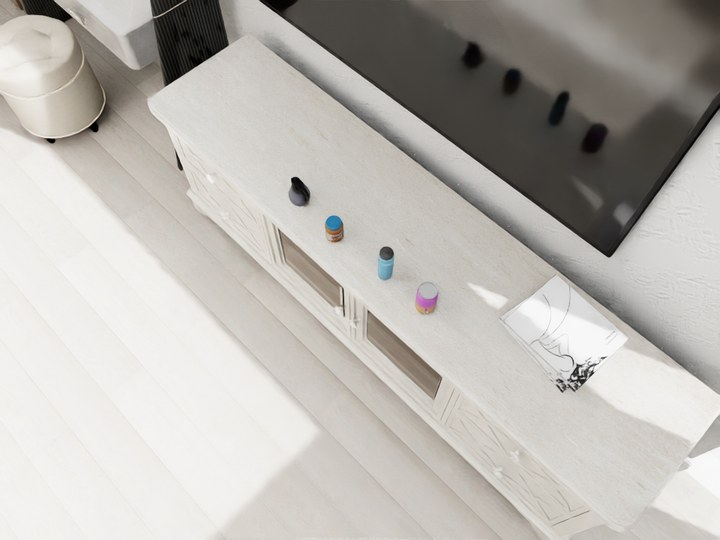}{4}{All four items are on the TV cabinet}\\\noalign{\vskip 8pt}
\extask{Object exchange}{two TidyBots; physical dependency: vacate, then fill}{Please move the Bowl-shaped sealed container from the smaller cabinet in the kitchen to where the green and white toy train was. Then move the green and white toy train from the dining table in the living room to where the Bowl-shaped sealed container was.}
\exframe{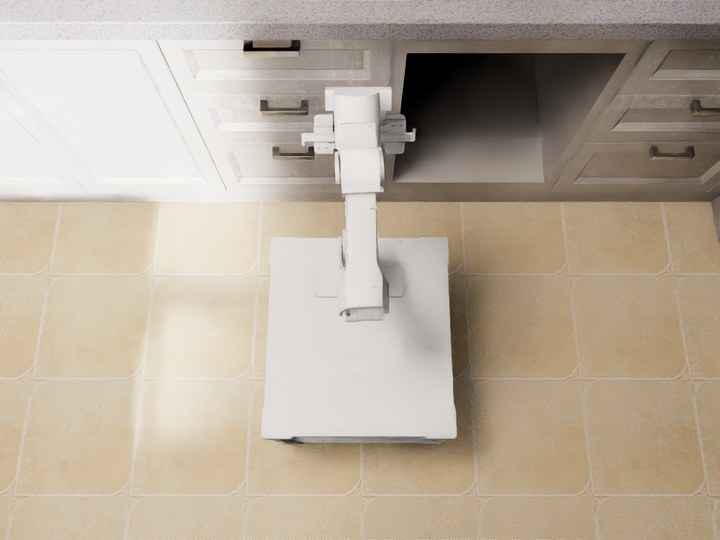}{1}{One TidyBot takes the container out of its place}
& \exframe{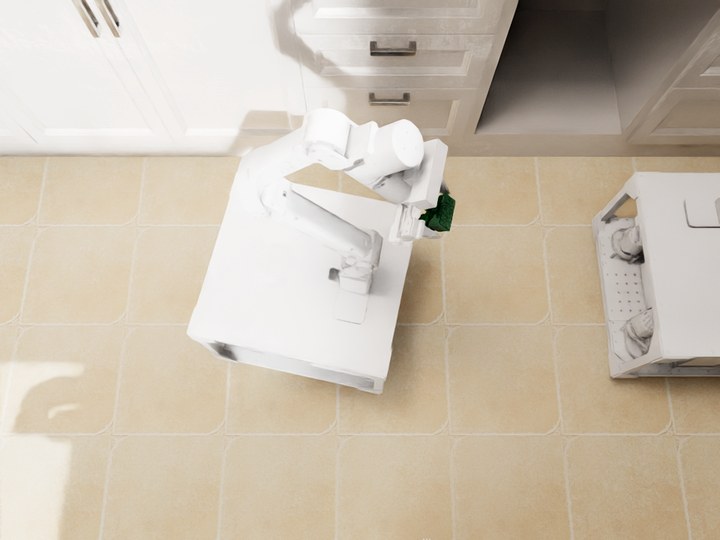}{2}{The other brings the toy train to where the container was}
& \exframe{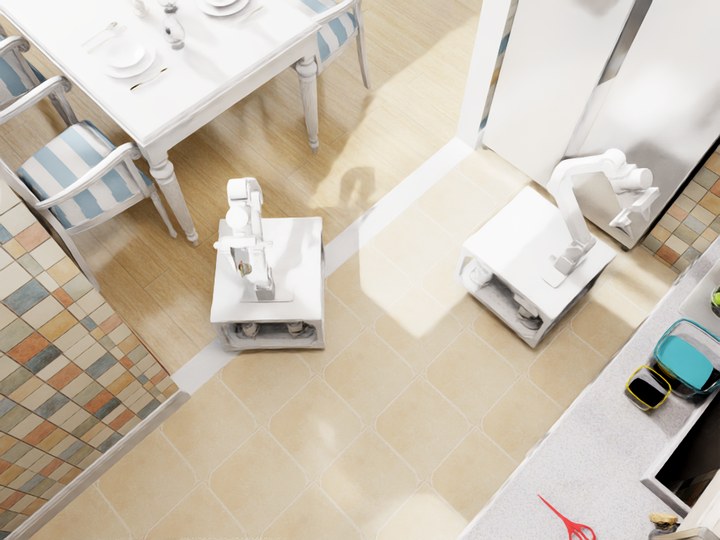}{3}{The container goes where the train was}
& \exframe{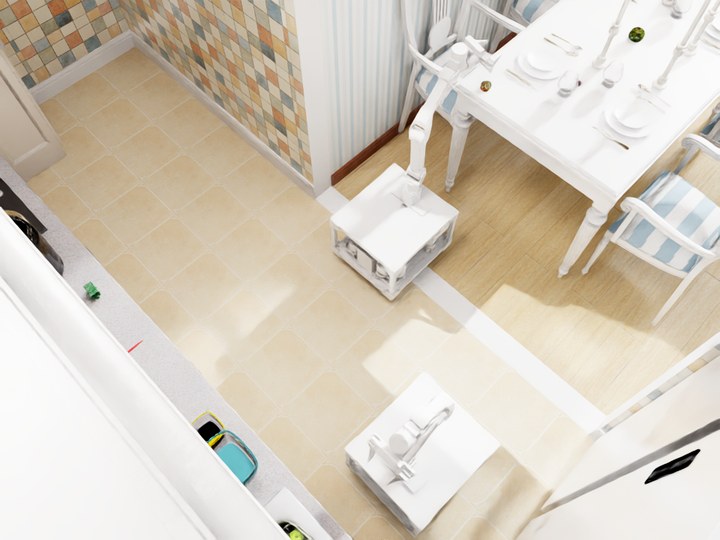}{4}{Both objects have swapped places}\\
\end{tabular}
\caption{Representative tasks with parallel workloads and an exchange. Each row gives the task family, the team, and the public instruction, followed by third-person renders of the joint Oracle reference execution recorded during qualification.}
\label{fig:task_examples}
\end{figure}

\begin{figure}[tb]
\centering\scriptsize
\setlength{\tabcolsep}{1.5pt}
\begin{tabular}{@{}llll@{}}
\extask{High-to-low relay}{R1 Pro and TidyBot; relay through a handoff surface}{Move the canola oil from the westernmost cabinet in the dining room onto the easternmost cabinet in the living room. Hand it off on the dining table. Leave it upright, released, and stable.}
\exframe{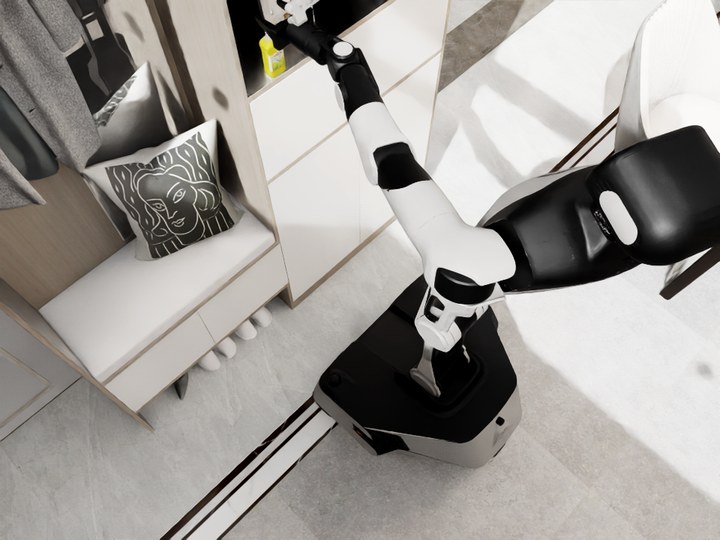}{1}{R1 Pro retrieves the oil from the high cabinet}
& \exframe{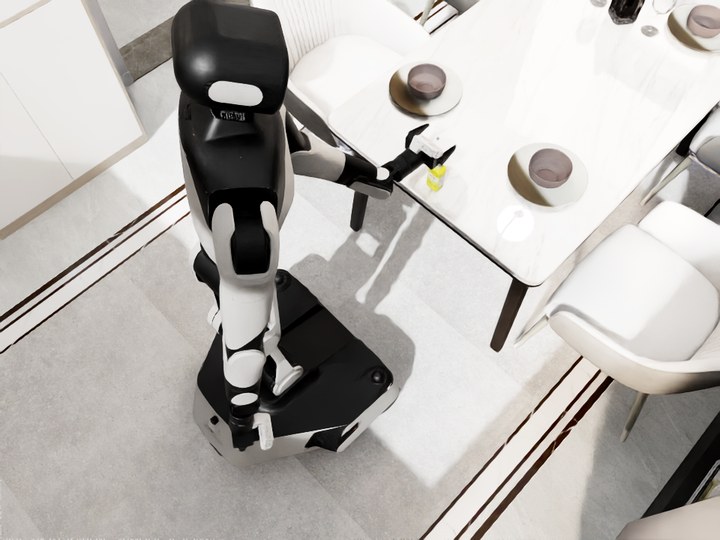}{2}{R1 Pro releases it on the dining table}
& \exframe{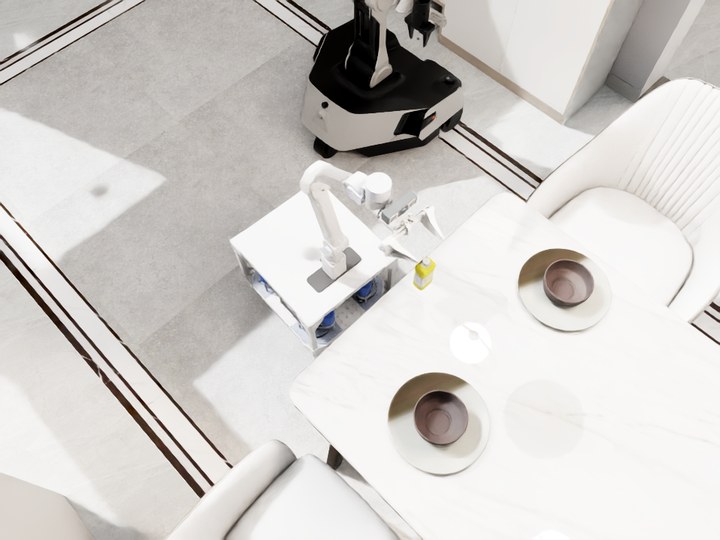}{3}{TidyBot grasps it from the handoff surface}
& \exframe{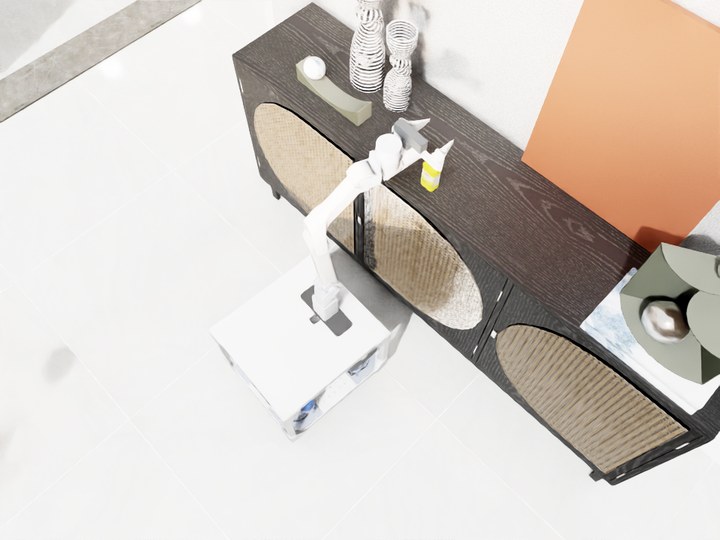}{4}{TidyBot places it on the target cabinet}\\\noalign{\vskip 8pt}
\extask{Narrow-passage relay}{R1 Pro and TidyBot; relay into a furnished aisle}{Move the condiment from the bookcase in the living room to the bedside table closest to the television in the living room at the end of the narrow aisle beside the wardrobes. Use the easternmost bedside table in the living room as the handoff surface. Leave all items upright, released, and stable.}
\exframe{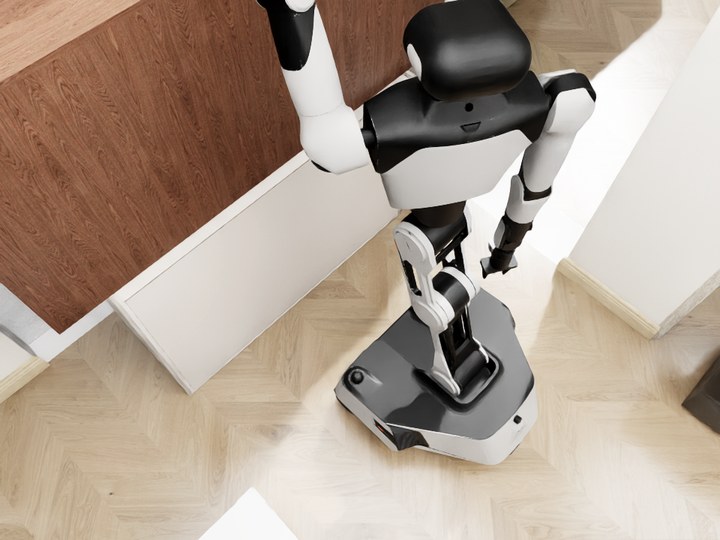}{1}{R1 Pro retrieves the condiment from the bookcase}
& \exframe{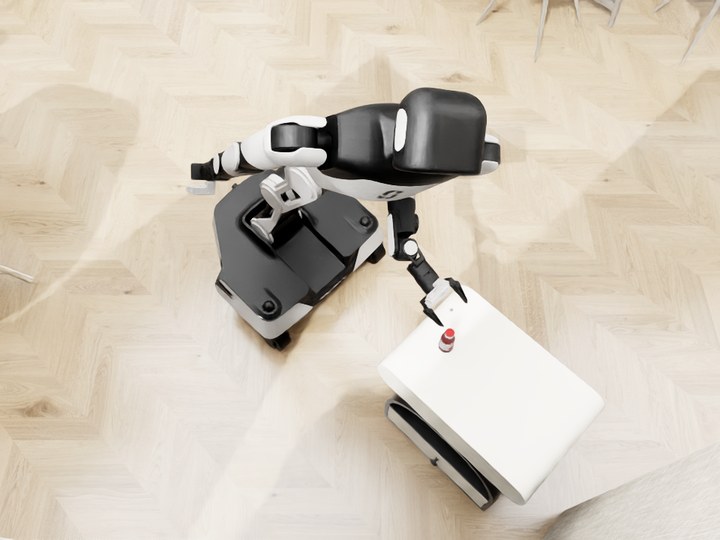}{2}{R1 Pro releases it on the handoff table}
& \exframe{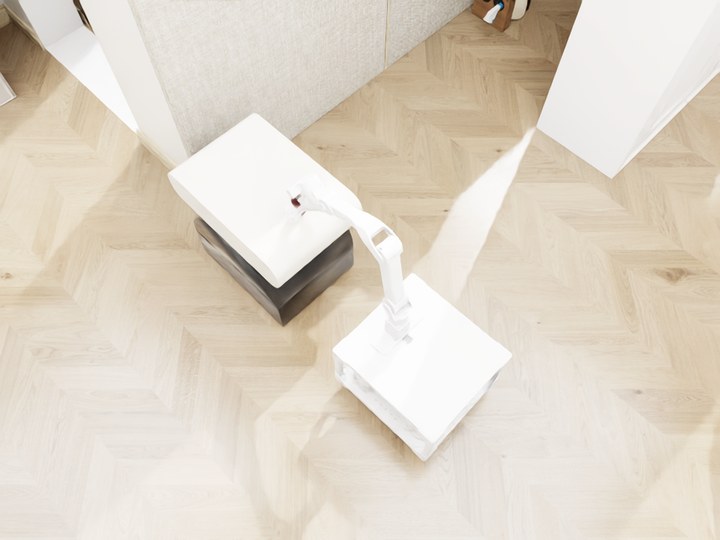}{3}{TidyBot takes it from the handoff table}
& \exframe{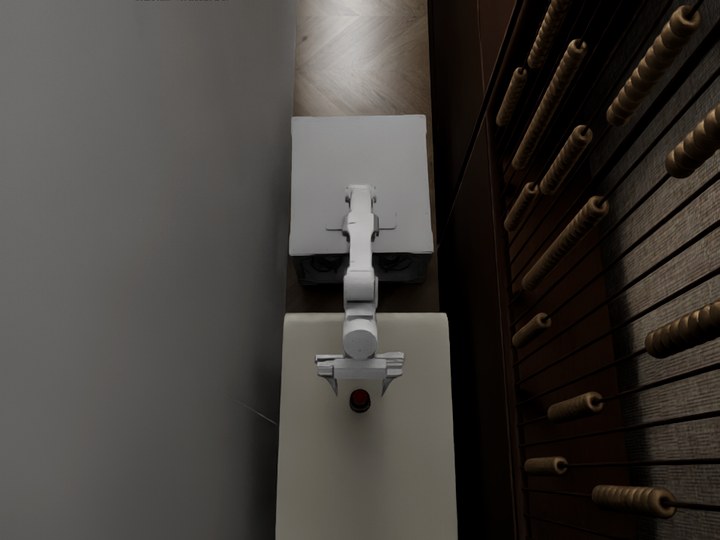}{4}{TidyBot delivers it inside the aisle}\\\noalign{\vskip 8pt}
\extask{Multi-object carrier transport}{R1 Pro and Create~3; loading and carrier transport}{Move the peanut butter jar and pitcher in a basket from the living-room dining table to the delivery point south of the table in the western bedroom. Leave both items released and stable in the basket at the delivery point.}
\exframe{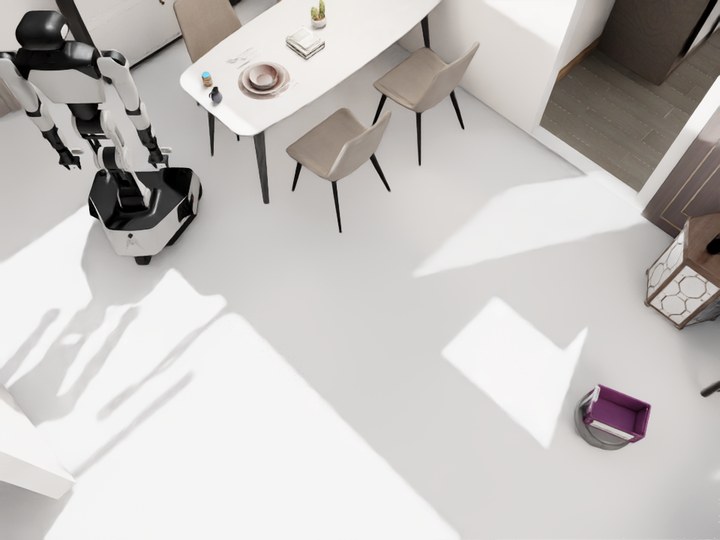}{1}{R1 Pro approaches the items; Create~3 waits}
& \exframe{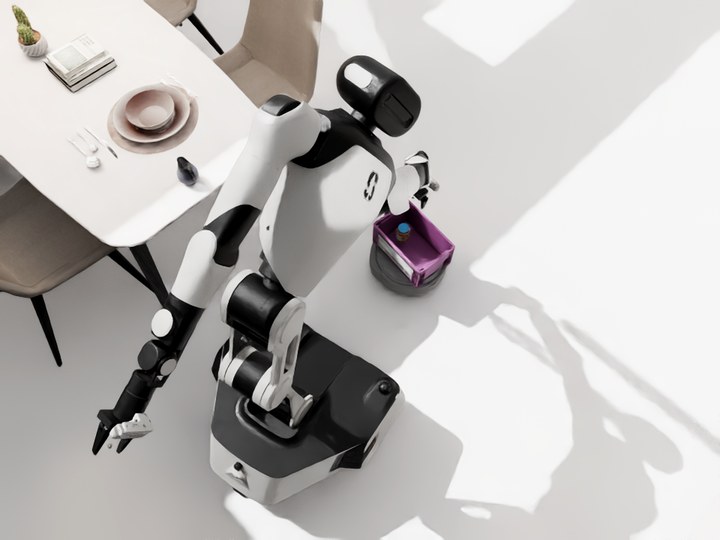}{2}{R1 Pro loads both items into the basket}
& \exframe{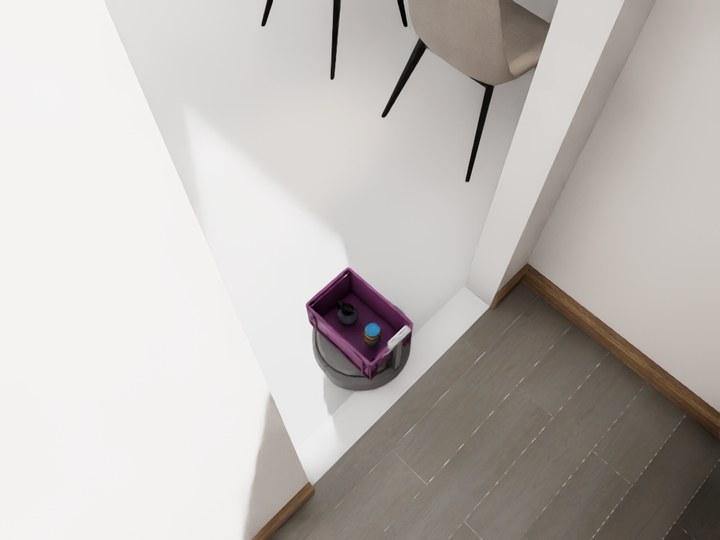}{3}{Create~3 carries them to the bedroom}
& \exframe{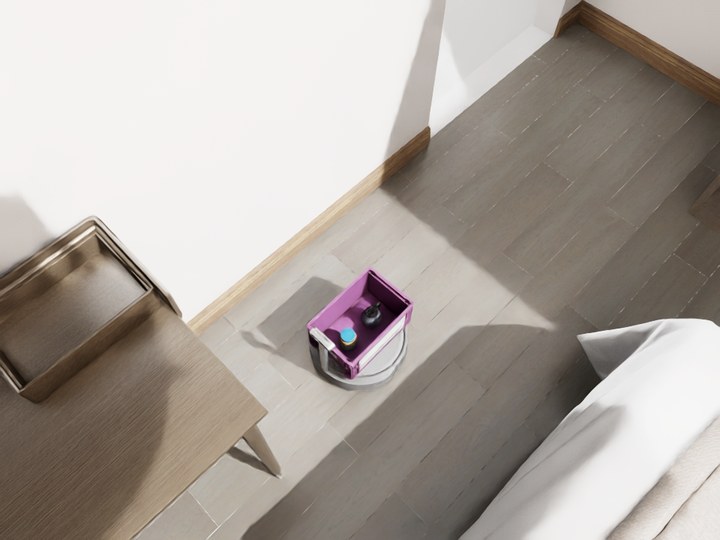}{4}{Create~3 stops at the delivery point}\\
\end{tabular}
\caption{Representative tasks with relays and carrier transport, shown as in \cref{fig:task_examples}.}
\label{fig:task_examples_relay}
\end{figure}

\Cref{fig:task_examples,fig:task_examples_relay} show six benchmark tasks, two with homogeneous and four with heterogeneous teams, covering parallel workloads, an exchange that requires vacating a location, relays through a handoff surface, and carrier transport. Each row gives the public instruction and four frames of the joint Oracle reference execution recorded during qualification. In the collection example, the Oracle reference stages the two robots' pickups one after the other.

\subsection{Family contracts and failure opportunities}
\paragraph{Homogeneous distribution, collection, packing, and restoration.}
Under the reference assignment $\alpha$ used for qualification, each robot handles distinct object--destination pairs (withheld during end-to-end evaluation); in restoration, one robot also closes a drawer after its own placement, without waiting for its teammate. A shared source creates competition for viewing, docking, or reaching space, and a shared destination can become occupied during placement.

\paragraph{Object exchange and clearance.}
One robot vacates a location that its teammate must fill or clears a route before the teammate's loaded traversal. The cleared state is evaluated when the consumer uses it, not when the producer announces completion. A moved obstacle may remain in the consumer's swept volume, or a teammate may re-occupy a cleared region.

\paragraph{High-to-low and low-to-high relays.}
The producer retrieves the object and releases it on an intermediate handoff surface for the receiver. Within the registered skill set, reaching horizontally over a cabinet or table from a collision-free floor stance with a top-down grasp posture bounds TidyBot--ARX5 capture TCPs at $z_{\mathrm{tcp}} \le z_{\mathrm{floor}} + 1.20$\,m relative to the scene floor datum. In all 18 high-to-low tasks, elevated cabinet sources place payload collision centers at $1.24\text{--}1.30$\,m above the floor, so every overhead capture TCP exceeds TidyBot's registered limit, whereas destinations lie on lower supports ($0.44\text{--}0.88$\,m) within TidyBot's workspace. Conversely, all 8 low-to-high tasks place sources on the floor ($0.05\text{--}0.08$\,m above $z_{\mathrm{floor}}$, below the $0.30$\,m low-source bound that excludes R1~Pro's registered torso/arm pickup profile) and final goals on elevated cabinets ($1.26\text{--}1.31$\,m, above TidyBot's $1.20$\,m limit). Qualification reconstructs actual capture TCPs from native fixed-joint attachments and URDF forward kinematics, and the evaluator checks physical release, support stability, ownership transition, and terminal placement.

\paragraph{Narrow-passage relay.}
Each of the 23 tasks places a 0.50\,m aisle along an existing wall of a VLNVerse home. A contiguous run of full-height wardrobes or cabinets, copied at original size from the scene, forms the other side; shallow cabinets or household items close any recess in the wall, and ordinary household items are scattered around the aisle. The bookcase sources place object centers at $1.63\text{--}1.67$\,m. Even under a zero-offset vertical link-length sum (ignoring joint-angle, horizontal-reach, and top-down posture constraints), the TidyBot--ARX5 URDF/USD chain bounds its TCP at 1.46\,m and its hand at 1.53\,m above the floor, so every source center lies above this vertical TCP bound by more than a 0.15\,m capture-assistance margin (after a 2\,mm settling allowance), and every source's collision bottom lies above the vertical hand bound. The destinations ($0.60\text{--}0.64$\,m) lie inside the aisle, whose 0.50\,m clear width admits the 0.40\,m-wide TidyBot base with 0.05\,m on each side but not the R1~Pro, whose base needs at least 0.60\,m even at $5^\circ$ tilt; its arms cannot pass over the full-height run, and the goal lies farther from the aisle mouth than the R1~Pro's base intrusion plus arm reach plus 0.15\,m. Qualification recomputes these four margins from the USD/URDF collision models, requires each to be strictly positive, and verifies them on the native trace together with TidyBot's loaded sweep through the aisle.

\paragraph{Carrier transport and two-ended loading/unloading.}
All 46 tasks (42 carrier transport, 4 two-ended) require transporting two objects in Create~3's KLT basket ($0.17\times 0.25\times 0.14$\,m inner volume). Create~3 (0.34\,m footprint) provides multi-item basket containment but has no manipulator, excluding solo pickup from elevated sources ($0.44\text{--}0.86$\,m) and solo unloading; R1~Pro and TidyBot have manipulators but carry no basket, excluding solo satisfaction of the basket-containment and carrier-transport predicates. Loading completes only after the payload settles in the basket, and transport succeeds only if every item remains supported until arrival. In the 4 two-ended tasks (three robots across rooms, ${\sim}10.6$\,m route), R1~Pro loads from an elevated bedroom table ($0.79\text{--}0.84$\,m), Create~3 transports the cargo across rooms, and TidyBot unloads each item onto a low living-room table ($0.45\text{--}0.49$\,m). Failures include missing items, premature departure, cargo ejection, and arrival poses outside unloading reach.

\begin{table}[tb]
\centering\small
\caption{Stratified cooperation-necessity and qualification evidence across all 500 benchmark tasks (350 homogeneous, 150 heterogeneous). Capability exclusions hold within the registered skill set, collision profiles, and task contracts (\cref{sec:qualification}).}
\label{tab:coop_evidence}
\setlength{\tabcolsep}{4pt}
\begin{tabularx}{\textwidth}{@{}>{\raggedright\arraybackslash}L{0.22\textwidth}c>{\raggedright\arraybackslash}L{0.14\textwidth}Y@{}}
\toprule
Stratum & Tasks & Team & Qualification check and single-robot exclusion within registered skills \\
\midrule
Homogeneous parallel-only & 241 & 2 Tidy / 2 R1 & Joint trace audit; occupancy-grid parallel-time model faster than lower bound over all legal solo orders from either start. \\
Homogeneous with dependency & 109 & 2 Tidy / 2 R1 & Same parallel-time filter plus trace-verified physical dependency establishment and consumer use. \\
\midrule
Height-asymmetric relays (18 high$\to$low, 8 low$\to$high) & 26 & R1 + TidyBot & High-source centers ($1.24\text{--}1.30$\,m) exceed TidyBot's horizontal-reach overhead capture limit ($z_{\mathrm{tcp}}\le z_{\mathrm{floor}}+1.20$\,m); floor sources ($0.05\text{--}0.08$\,m $<0.30$\,m) exclude R1~Pro's registered pickup profile, paired with $1.26\text{--}1.31$\,m goals; URDF-FK trace audit of attachment TCPs and handoff ownership. \\
Narrow-passage relays & 23 & R1 + TidyBot & Bookcase source centers ($1.63\text{--}1.67$\,m) exceed even TidyBot's zero-offset vertical link-sum TCP bound (1.46\,m) by more than a 0.15\,m assistance margin, and source bottoms exceed its vertical hand bound (1.53\,m); the 0.50\,m furnished aisle admits the 0.40\,m-wide TidyBot base but not R1~Pro ($\ge 0.60$\,m at $\le 5^\circ$ tilt), whose arm reach from the mouth falls short of the goal; four positive margins verified on native traces. \\
Carrier \& two-ended transport (42 carrier, 4 two-ended) & 46 & R1 + Create~3 (+ TidyBot) & Two-item basket contract: Create~3 (0.34\,m footprint) has the basket but no arm to pick ($0.44\text{--}0.86$\,m) or unload; R1~Pro and TidyBot have arms but no basket. Two-ended tasks bind R1 elevated loading ($0.79\text{--}0.84$\,m), Create~3 inter-room transit, and TidyBot low-table unloading ($0.45\text{--}0.49$\,m). \\
Clear-then-deliver \& two-source collection (25 + 30) & 55 & R1 + TidyBot & Distinct workstation profiles (R1~Pro reaches elevated supports; TidyBot's registered capture TCP is limited to $z_{\mathrm{floor}}+1.20$\,m): 44/55 pair elevated R1 cabinets ($1.13\text{--}1.25$\,m, overhead TCP $>1.20$\,m or body-clutter constrained) with compact TidyBot stances ($0.38\text{--}0.83$\,m), and 11/55 bind furniture-gap TidyBot stances with deeper R1 tabletop reaches outside peer endpoint tubes. All 55 pass an exact DP solo-cost lower bound over all legal serial orders ($\ge 2\times$ bound using measured action durations and straight-line chords), and all 25 clearance tasks verify cooked-hull region vacancy before delivery. \\
\bottomrule
\end{tabularx}
\end{table}

\paragraph{Clear then deliver and two-source collection.}
All 55 tasks (25 clear then deliver, 30 two-source collection) assign four objects across three distinct furniture supports (two objects per robot; eight pick-and-place actions). In clear then deliver, TidyBot removes two occupying objects from a shared dining table ($0.77\text{--}0.82$\,m) to a low cabinet ($0.58\text{--}0.62$\,m) before R1~Pro delivers two incoming objects from a tall cabinet ($1.14\text{--}1.20$\,m) onto the cleared region. Two-source collection gathers two objects per source onto one common receiving surface, counting each object once. Both families combine embodiment-conditioned workstation complementarity with an exhaustive solo-schedule lower bound (\cref{tab:coop_evidence}).

\subsection{Qualification details}
\label{supp:qualification}
A reference execution uses the same robot and object models as evaluation. Scene geometry, gravity, collision approximations, joint limits, gripper rules, and payload constraints are frozen with the record. The Oracle may plan with privileged information but cannot set target states directly, and an independent check replays the state and event traces rather than relying on controller flags. Heterogeneous candidates run once without retries, and an independent audit re-derives contacts, carrier containment, cooked-hull region clearance, capture-TCP heights, and terminal stability (75 samples below 2\,cm/s) from the trace.

\Cref{tab:coop_evidence} details the qualification evidence by stratum across all 350 homogeneous and 150 heterogeneous tasks ($26 + 23 + 46 + 55 = 150$). For homogeneous tasks, the time model moves each robot at uniform speed along occupancy-grid routes and includes joint-schedule waits and clearance motions, while the single-robot baseline minimizes over either start and all precedence-legal orders. For the 55 heterogeneous clear-then-deliver and two-source collection tasks, qualification enforces two layers: (i)~embodiment-specific workstation conditioning, where candidate stances are filtered under each robot's collision cylinder, vertical reach, and peer endpoint-separation tubes of radius $r_{\mathrm{R1}}+r_{\mathrm{Tidy}}+2\epsilon$, where $r$ is each robot's planning radius, excluding a single embodiment from covering both the elevated or deep tabletop stances and the tight low-furniture stances within the registered workstation map; and (ii)~an exact dynamic-programming lower bound over all precedence-legal serial one-payload orders for either single actor, which retains measured pick/place durations from the joint reference and charges missing inter-station transfers at straight-line chord distance divided by each robot's native speed cap (favoring the single robot, with turns and preparation relaxed to zero), requiring at least a $2\times$ registered-domain makespan advantage alongside a non-interfering native reference check. Across all four heterogeneous strata, Oracle qualification thus verifies both continuous joint feasibility and the necessity of combining complementary robot capabilities within the registered skill set.

\subsection{Stable final-state predicates}
A placement predicate specifies object identity, target region, physical support or containment, release from the gripper where required, and a stability interval of 0.5\,s. Whole-object containment is used when the task requires the object to fit, rather than testing only its center. Thresholds for translational and angular speed, contact tolerance, and overhang are fixed in the released evaluator configuration and were set before any baseline was evaluated. Regression tests cover boundary-straddling objects, momentary contact, wrong-object placement, late payload loss, and goals invalidated by the teammate.

\FloatBarrier
\section{Information Boundary and Conformance}
\label{supp:information}
\subsection{Sensor data versus robot metadata}
Privileged Oracle access to scene state is restricted to offline task generation and qualification and to the explicitly labeled diagnostic controls in \cref{sec:diagnostic} ($s_d^-$ initialization, reference-producer, and Oracle-guided trials), and is never exposed to policies in end-to-end or unguided runs. At evaluation time, the policy contract specifies the public camera names, resolution, field of view, RGB/depth synchronization, metric-depth convention, invalid-depth encoding, timestamp source, sensing rate, and declared sensor model.

Proprioception consists of explicitly listed measured quantities, such as articulated-joint positions/velocities and gripper opening. Base velocity is accepted only with its source stated. A simulator's world-frame linear velocity, or its virtual prismatic base joints, must not enter under the generic name ``proprioception'' without being recognized as assistance. IMU acceleration and angular velocity do not provide exact global position. Any orientation estimate has a declared reference frame and estimation/noise assumptions.

Public robot metadata includes static camera-to-link extrinsics, kinematic parameters, joint limits, robot geometry, gripper dimensions, controller gains or semantics required to issue valid commands, and the control period. This is distinct from access to furniture meshes and object models in the current scene. For a wrist camera, the base-relative camera transform is computed from measured joints and fixed hand--eye calibration. Neither that transform nor a self-collision model requires an oracle map of the environment.

\subsection{Localization, maps, and cross-robot frames}
The primary comparison uses the same sensor-derived odometry implementation for all configurations requiring persistent metric coordinates. Its model/checkpoint, input modalities, initialization, update rate, and uncertainty interface are recorded. If this implementation is not available, the result must be labeled as local-frame control or as a declared localization-assisted setting; exact simulator-relative pose cannot be substituted without changing the track.

Each actor's local odometry begins in its own arbitrary frame. Shared observations and messages do not create a known transform between these frames. Alignment must be estimated from permitted observations, or an initial relative transform must be listed as explicit assistance for every method. Shared-workspace messages may use semantic regions without claiming exact global geometry. A preprovided room layout and static occupancy map constitutes the declared scene prior (used by the +prior baselines and \method in \cref{tab:main} and by \method in \cref{tab:diag}(b)--(d), without task-object positions or exact robot localization) and is withheld in the default no-prior track.

Environment occupancy and collision geometry are reconstructed from visible depth and admissible history. Previously observed geometry may persist in memory; unseen regions remain unknown according to a fixed planner policy. Dynamic changes, including teammate motion and moved objects, update the map through permitted observations or messages. A teammate may communicate its estimated state and known body geometry, but the common runtime must not attach its hidden exact world pose to that message.

\subsection{Tools and feedback}
Object localization, segmentation, grasp proposal, and progress verification may use pretrained models, but their online inputs must satisfy the public contract. Model identity, checkpoints, and any task-specific training are recorded. A tool may return its own planning failure, controller tracking error, or visual confidence, but never an evaluator predicate. ``The gripper commanded closed'' and ``the intended object is securely held'' are distinct observations.

A ground-truth terminal score is used for evaluation. In the main test loop, the policy cannot repeatedly query that score or a partial-goal checklist. An episode termination signal, if required by the environment, ends the rollout immediately; it cannot be queried and then used to continue revising the same episode. Richer training feedback, reset-enabled adaptation, and oracle diagnostics are distinct protocol dimensions with separate logs.

\begin{table}[tb]
\centering\small
\caption{Information contract of the default no-prior sensor-grounded track, together with declared priors and diagnostic inputs (\cref{sec:execution,sec:diagnostic}). ``Derived'' means computed exclusively from permitted data.}
\label{tab:information}
\begin{tabularx}{\textwidth}{@{}L{0.24\textwidth}L{0.18\textwidth}Y@{}}
\toprule
Quantity or capability & Track / status & Conformance criterion \\
\midrule
RGB-D, joints, gripper, IMU & Public sensors & Fixed modalities, frames, rates, timestamp and noise semantics \\
Robot URDF / own collision mesh & Public metadata & Contains the robot, not task-scene assets or hidden target geometry \\
Camera-to-base transform & Derived / calibrated & Fixed extrinsic plus FK; no simulator world-pose query \\
Persistent base localization & Derived & Sensor estimator with logged inputs; no exact pose renamed ``odom'' \\
Teammate position & Derived / communicated & No undeclared common world transform \\
Object masks and poses & Derived & Learned perception or geometric estimation from observations \\
Scene map / collision model & Derived (no-prior) & Built from allowed observations and communication \\
Static scene prior (+prior) & Declared prior & Room layout and static occupancy from scene assets, without task-object poses or exact localization (used by +prior baselines and \method in \cref{tab:main}, and by \method in \cref{tab:diag}(b)--(d)) \\
IK / self-collision planning & Allowed & Uses robot model and measured state \\
Task success / hidden predicates & Evaluator only & No online oracle query or feedback-driven test restart \\
Oracle poses, grasps, waypoints & Oracle diagnostic & Object 6D poses, grasp states, and navigation waypoints; withheld in all end-to-end and unguided runs, supplied only in \cref{tab:diag}(d) \\
Reference snapshots & Diagnostics only & Full-state restoration declared; no continuous-score substitution \\
\bottomrule
\end{tabularx}
\end{table}

\subsection{Isolation and audit tests}
The deployment separates the simulator/evaluator process from agent and tool clients. The client exposes a typed public schema rather than a native environment reference. The policy workspace must not mount task-instance source files, evaluator logs, reference trajectories, hidden scene assets, or a simulator object registry. Error messages and visualization exports are filtered through the same boundary. Import restrictions alone are insufficient, since an injected object may still reference a privileged host.

An audit records every policy-facing field's provenance and every planner input's construction path. Representative tests include: denying direct environment-object access; checking that camera-to-base transforms match FK; verifying that movement changes odometry through the estimator; testing that unknown geometry is not treated as an oracle free-space query; and confirming that train/test feedback permissions differ as specified.

A noninterference test evaluates a pure planning or observation tool twice with identical permitted history and random seed while changing a hidden, unobserved object in a shadow world; the tool's output must stay unchanged. The test detects policy-side access to information outside the declared inputs, whereas later physical outcomes may differ once a robot encounters the changed object.

Audit outcomes and test coverage are released with the benchmark. A failed conformance check invalidates the sensor-grounded label until it is fixed.

\section{Methods and Adaptations}
\label{supp:adapters}

\subsection{Shared worker and comparison units}
Each evaluated configuration is specified by its language-model endpoint and reasoning setting, prompt and memory format, agent runner, declared prior information, frozen development artifacts, communication access, and episode budgets. Our baseline adaptations follow the public implementations of ASPIRE~1.0~\citep{aspire,aspirecode} and RAX~\citep{rax,raxcode}. Within each comparison of \cref{tab:main,tab:diag}, \method, the sensor-grounded baselines (ASPIRE-SG and RAX-SG), and their scene-prior variants (+prior) execute on the same frozen worker version---\ie, the same physics simulation, sensor-grounded perception and control tools, odometry and occupancy estimator, team communication primitives, and evaluator. Differences across rows therefore reflect agent organization, declared scene priors, and development artifacts (\cref{sec:methods}).

\subsection{ASPIRE-SG}
The adaptation retains ASPIRE's tool registry, program executor, program validation, and execution-feedback-driven program revision~\citep{aspire}. In the inspected BEHAVIOR implementation, odometry is computed from exact simulator poses, navigation can query the scene traversability map and shortest paths, and motion planning accesses the simulator collision world~\citep{aspirecode}. These are replaced by the shared estimator and observation-derived geometry, including target-exclusion logic that would otherwise use hidden object bounding boxes. One ASPIRE agent runs per robot; the agents receive the shared instruction, their own observations, the team board, and teammate messages, and assign work among themselves. Reference roles are supplied only in the diagnostic. Upstream ASPIRE accumulates skills over single-robot episodes with simulator success feedback; as the baselines have no development phase (\cref{sec:methods}), this exploration is disabled before testing and the agents instead access a shared library of generic, non-task-specific stage templates.

\subsection{RAX-SG}
The adaptation retains RAX's MCP context, \texttt{main(ctx)} program runner, and snapshot verification~\citep{rax,raxcode}. Its BEHAVIOR interface is already close to the sensor-grounded contract: RGB-D, named proprioception, base-relative camera calibration, integrated body-velocity odometry, and robot-only motion preview~\citep{raxcode}. The common estimator and metadata replace backend-specific quantities where needed. One RAX agent runs per robot with the same team board, messages, and stage-template library as ASPIRE-SG. RAX's training-time exploration and code accumulation are not run, optional demonstration images and richer goal descriptions are not supplied, and episode reviews based on simulator goal reports are disabled.

\paragraph{Program validation and revision under heterogeneous coordination.}
Following upstream ASPIRE and RAX, each baseline turn must contain either one executable multi-step program (\texttt{code}, up to four tool calls) or one atomic action object. A reply that violates this contract, for example by concatenating unsequenced message and motion objects or by placing JSON task-dispatch lists in the \texttt{code} field, returns structured validation feedback and is revised within the episode; revisions consume shared-clock time (\cref{sec:execution}). \method instead separates task-graph negotiation from single-step, schema-validated tool calls.

\subsection{\method}
\label{supp:comoma}

\paragraph{Team execution.}
A team session owns the simulation clock. In each control cycle it submits the current targets of all team members (two robots in 98 test tasks, three in the two two-ended tasks) and advances four physics steps at 120\,Hz. Planning requests run in the background; idle, stale, or cancelled actors hold their measured pose. The runner follows ASPIRE's design of a tool registry and a program executor~\citep{aspirecode}; the task graph, team negotiation, room graph, object beliefs, and asynchronous skill wiring are our implementation.

\paragraph{Task graph and coordination layer.}
Nodes encode executable subtasks with the actor that the agent assigns, required capabilities, and prerequisites. A node becomes eligible after its prerequisite completion reports are accepted, and completion requires observation-grounded evidence. Operations on a shared object, workspace, or handoff surface request a lease; claims, acknowledgements, and completion reports carry episode and node identities. The task graph and lease rules stay fixed during development.

\begin{figure}[tb]
\centering
\resizebox{\linewidth}{!}{\begin{tikzpicture}[
  font=\sffamily\scriptsize, >={Stealth[length=4pt,width=3.2pt]},
  every node/.append style={execute at begin node={\hyphenpenalty=10000\exhyphenpenalty=10000}},
  box/.style={draw=#1!55!black, fill=#1!5, rounded corners=2pt, line width=0.55pt,
    align=center, inner sep=3pt, minimum height=1.0cm, text width=2.45cm},
  box/.default=gray,
  metric/.style={draw=teal!60!black, fill=teal!6, rounded corners=2pt, line width=0.55pt,
    align=center, minimum height=0.6cm, text width=1.3cm, inner sep=1.5pt},
  arr/.style={->, line width=0.6pt, draw=black!70},
  lab/.style={font=\sffamily\tiny, text=black!65, inner sep=1.5pt},
  panel/.style={draw=black!28, rounded corners=4pt, line width=0.5pt},
  title/.style={font=\sffamily\footnotesize\bfseries, anchor=north west, inner sep=0pt}
]
\draw[panel] (0,0) rectangle (10.15,5.0);
\draw[panel] (10.95,0) rectangle (16.9,5.0);
\node[title] at (0.2,4.8) {(a) Development on withheld seeds};
\node[title] at (11.15,4.8) {(b) Evaluation on held-out seeds};

\node[box=blue]   (lib)  at (1.6,3.15) {\textbf{Skill library}\\ + navigation memory};
\node[box]        (run)  at (5.05,3.15) {\textbf{Sensor-grounded}\\ team rollouts};
\node[box]        (evi)  at (8.5,3.15) {\textbf{Failure evidence}\\ instruction, egocentric\\ frames, tool receipts};
\node[box=violet] (dia)  at (8.5,1.3)  {\textbf{Diagnostic model}\\ proposes a skill revision};
\node[box=green]  (gate) at (5.05,1.3)  {\textbf{Retention gate}\\ unit/regression tests,\\ fresh native rerun};

\draw[arr] (lib) -- (run);
\draw[arr] (run) -- (evi);
\draw[arr] (evi) -- (dia);
\draw[arr] (dia) -- (gate);
\draw[arr] (gate.west) -| node[lab, pos=0.28, above] {pass: retain} (lib.south);
\draw[arr, densely dashed, draw=black!40] (gate.south) -- ++(0,-0.42) node[lab, right, pos=0.6] {fail: discard};

\draw[densely dashed, line width=0.6pt, draw=black!50] (10.55,0.15) -- (10.55,2.2);
\draw[densely dashed, line width=0.6pt, draw=black!50] (10.55,3.2) -- (10.55,3.9);
\begin{scope}[shift={(10.55,2.75)}]
  \draw[line width=0.7pt, draw=black!70] (-0.11,0.02) -- (-0.11,0.16) arc[start angle=180, end angle=0, radius=0.11] -- (0.11,0.02);
  \fill[black!70, rounded corners=0.5pt] (-0.17,-0.2) rectangle (0.17,0.04);
\end{scope}
\node[lab, font=\sffamily\tiny\bfseries] at (10.55,2.38) {freeze};

\draw[arr, draw=blue!60!black, line width=0.75pt]
  (lib.north) -- ++(0,0.6) -| node[lab, pos=0.27, above, text=blue!50!black]
  {frozen artifacts, loaded by every team member} (12.3,3.65);

\node[box=blue, text width=2.25cm]   (frz) at (12.3,3.15) {\textbf{Frozen CoMoMa}};
\node[box=orange, text width=2.7cm] (bas) at (15.3,3.15) {\textbf{Adapted baselines}\\ ASPIRE-SG, RAX-SG\\ ($\pm$ scene prior)};
\node[box, text width=5.3cm, minimum height=0.8cm] (tst) at (13.95,1.95)
  {100 test tasks, one held-out seed each\\ same sensing, tools, and budgets};
\draw[arr] (frz.south) -- (frz.south |- tst.north);
\draw[arr] (bas.south) -- (bas.south |- tst.north);

\foreach \x/\t [count=\i] in {11.7/{SR, GC}, 13.2/{NSR, PE}, 14.7/{HCov, HSR}, 16.2/{Calls,\\ tokens}} {
  \node[metric] (m\i) at (\x,0.85) {\t};
  \draw[arr] (\x,1.55) -- (m\i.north);
}
\end{tikzpicture}}
\caption{\method development and evaluation. On the withheld seeds, failure evidence from development episodes drives candidate skill revisions, which are retained only after software tests and a fresh native rerun. The skill library and navigation memory are then frozen and evaluated on one held-out seed per test task, with the same sensing, tools, and budgets as the baselines.}
\label{fig:learning_protocol}
\end{figure}

\paragraph{Development.}
\method's skill library was built with an offline repair loop on the withheld development seeds, which are disjoint from the evaluation seeds (\cref{fig:learning_protocol}). From sensor-grounded development rollouts---recording the public instruction, egocentric camera frames, and tool execution receipts---a diagnostic model identifies failure modes in generic skill preconditions or recovery logic and proposes candidate revisions. A candidate skill revision is retained only if it passes unit and regression tests and succeeds on a fresh native rerun on the withheld seed. Skills encode embodiment- and mechanism-level control logic rather than task-specific data. The navigation memory compiles topological route graphs from the same rollouts; at test time it proposes subgoals, while collision avoidance, base planning, and grasping use current observations. Both artifacts are frozen before evaluation, and all team members load the same versions and ground them in their own observations.

\subsection{Budgets}
\label{supp:budgets}

\begin{table}[tb]
\centering\small
\caption{Measured model use per end-to-end episode (means over 100 episodes). Calls are averaged per robot across the 2--3 team members, and tokens are summed per episode, on homogeneous (H) and heterogeneous (X) tasks.}
\label{tab:budgets}
\setlength{\tabcolsep}{3pt}
\begin{tabular}{@{}lcc@{}}
\toprule
 & Calls used & Tokens ($10^3$) \\
Method & H / X & H / X \\
\midrule
ASPIRE-SG & 9.5 / 12.0 & 533 / 88 \\
RAX-SG & 10.4 / 11.0 & 591 / 83 \\
ASPIRE-SG + prior & 9.4 / 12.0 & 592 / 125 \\
RAX-SG + prior & 9.4 / 11.5 & 595 / 118 \\
\method & 12.4 / 24.4 & 1401 / 523 \\
\midrule
\method (GPT-6.1-sol) & 9.9 / 19.1 & 1341 / 831 \\
\bottomrule
\end{tabular}
\end{table}

\Cref{tab:budgets} reports the measured model use. All rows except the last use GPT-5.6-luna with medium reasoning effort and the same 20\,min of model time and 500\,s of simulated time per episode; programs of ASPIRE-SG and RAX-SG contain up to four tool calls, whereas each \method decision issues one. \res{Each model call returns at most 8{,}192 output tokens, and the per-episode budget of model time excludes provider-side rate-limit waits.}

\subsection{Methods not included}
CaP-Agent0~\citep{capx,capxcode} regenerates a program per instance with test-time retries and would require a separate reset-enabled protocol. EMOS~\citep{emos} is a relevant multi-robot system, but its low-level skills use simulator-provided object states (\cref{tab:comparison}), so a sensor-grounded adaptation would replace most of its execution stack.

\section{Execution and Scoring Details}
\label{supp:evaluation}

\subsection{Shared physical timeline}
A physics step advances all robots and dynamic objects together at 120\,Hz, and robot targets are committed every four steps. Model requests run asynchronously and the physics does not pause for them; a robot without a new target holds its measured pose. A long blocking skill of one robot therefore never suspends its teammate. All methods use the same model endpoint and simulator, so simulated-time budgets are comparable. Model calls and tokens per episode are summarized in \cref{tab:budgets}.

\subsection{Stopping, recovery, and incidents}
An episode stops when all goals hold, the simulated-time or inference budget is exhausted, or a task-defined unrecoverable violation occurs. A local failure of one robot does not terminate its teammate unless a required dependency prevents continuation. The evaluator can record a completed goal and later remove it if a subsequent action invalidates its terminal condition.

Test episodes have no reset privilege. Grasp retries, replanning, and program edits within the same evolving scene are permitted recovery actions; restoring an earlier simulator state or rerunning a seed after observing it is not. An infrastructure incident, meaning a host, service, or simulator failure unrelated to any policy decision, receives a logged rerun with the same seed and no transfer of policy knowledge. Policy-caused invalid commands, planner exhaustion, or failure to recover before the deadline remain failures. Incidents and reruns are logged with each episode.

\subsection{Metric details}
\label{supp:metric_details}
\paragraph{Goal completion.}
GC is computed per episode as the fraction of task checkpoints completed at termination and then averaged over episodes. Checkpoints follow the task program and cover both navigation and manipulation. A navigation checkpoint credits the first arm-equipped robot that brings its base within $r=1.0$\,m of an object before the object's first acquisition, or the carrier that comes within $r$ of the loading robot. Manipulation checkpoints cover each object's acquisition, transport, and placement, the carrier's payload loading and arrival, and fixture goals such as an articulated joint in restoration tasks. A navigation checkpoint whose occupancy-grid path from the robot's pose at reset is shorter than 0.5\,m is already satisfied at reset and is excluded. A checkpoint counts only once its physical criteria are verified from object and robot state.

\paragraph{Navigation legs.}
Each navigation checkpoint defines one leg. Because agents allocate work themselves, a leg is attributed to the robot that first reaches its target or, if no robot reaches it, to the arm-equipped robot that comes closest; it starts at that robot's initial pose or at its previous arrival. A leg is reached when this robot's base comes within $r$ of the target. The measured length $L_j$ sums the planar displacement of the base over the leg from simulator samples, including retries and corrective motions. The reference length $L_j^{\rm ref}$ is the shortest 8-connected path on the scene's 2D occupancy grid (0.05\,m cells) from the leg's start to the free cells within $r$ of the target, computed with A*, so it follows the team's own allocation rather than the Oracle's. All reported NSR and PE values use $r=1.0$\,m.

\paragraph{Pre-handoff states and handoff predicates.}
The pre-handoff state is reached when the producer holds the dependency object next to the handoff surface (relays), holds a payload object within reach of the carrier's basket (carrier loading), or holds the occupying object at the cleared region (clearance). Establishment requires a stable release on the handoff surface within the receiver's reachable region, a settled payload inside the basket, or a vacated region, respectively, still holding when the consumer starts. Valid use requires the robot that starts using the condition to grasp the same object, depart and arrive with the payload supported, or complete its placement or traversal through the cleared region. Agents allocate work themselves, so every stage is attributed to the robot that physically performs it: a single robot holding the object, or for a grasp not yet closed, a single robot in contact with it. Multiple robots on one object earn no credit, and a takeover discards the first robot's partial evidence.

\paragraph{Dependency accounting.}
HCov keeps every dependency of $\mathcal D$ in its denominator, including those whose producer was never reached. HSR uses reached dependencies whose condition a robot other than the producer starts to use; a dependency that the producing robot resolves itself is counted as self-resolved and excluded from HSR. A method that avoids difficult dependencies can therefore show a high HSR but a low HCov, and both are reported. Strict team success requires all final goals, all required dependencies, and all mandatory constraints.

\subsection{Statistical reporting}
The unit of analysis is the parent task. Each method runs one evaluation seed per test task, and every rate is reported together with its denominator. End-to-end differences between methods are paired by task and summarized with a 95\% bootstrap confidence interval over tasks (10{,}000 resamples). In the handoff-isolated diagnostic, methods are compared on the same dependencies, and paired differences are tested with an exact McNemar test.

\section{Handoff-Isolated Diagnostic and Failure Attribution}
\label{supp:diagnostics}

\subsection{Constructing pre-handoff states}
Following reference-initialized skill evaluation~\citep{maniunit}, each diagnostic trial starts from the pre-handoff state $s_d^-$ of one required dependency on an evaluation seed. We obtain $s_d^-$ by executing the joint Oracle reference of that seed up to the producer's establishing action; the state is reached through real simulated actions rather than by assigning object poses. Before a trial is counted, a consistency check verifies object identity, robot base and joint states, gripper and grasp state, object pose and support, and the teammate configuration against the reference. States that fail the check are excluded and counted. The evaluated method receives the same rendered observations as in a continuous episode, the public instruction, and a summary of completed subtasks, but no reference actions.

\subsection{Conditions}
Both conditions share the task contract, the local budget of \res{20}\,min of model time, and the success definition.
\begin{itemize}[leftmargin=*,itemsep=1pt]
    \item \textbf{Full}: the evaluated method controls both producer and consumer from $s_d^-$.
    \item \textbf{Reference producer}: the Oracle executes the producer's establishing action physically; the evaluated method then controls all team members from the resulting state.
\end{itemize}
Comparing the two conditions separates failures in establishing the produced condition ($E_d$) from failures in using it ($U_d$). CKPT counts the task checkpoints completed during the trial over all checkpoints of the task; checkpoints already satisfied at $s_d^-$ are not credited. The diagnostic instantiates 107 of the 114 test dependencies (omitting the second clearance in seven object-exchange tasks), of which \res{98} pass the pre-handoff physical consistency check identically across all methods and backbones.

\subsection{Failure attribution criteria}
Post-hoc failure analysis (\cref{sec:failures}) attributes rollouts using the diagnostic criteria in \cref{tab:failure_taxonomy} from recorded tool receipts, motion-planner and IK feedback, team messages, and evaluator state traces. Each failed rollout or stage is attributed to its first substantive failure rather than downstream consequences: an unreached source is recorded under navigation rather than acquisition, and failed point-grasp requests are separated by motion-execution rejection (inverse kinematics or collision-rejected arm paths) versus post-grasp loss. Workspace contention is recorded only when a time-aligned trace links the teammate's occupancy to an attempted motion or grasp; waiting alone is not contention.

\begin{table}[tb]
\centering\small
\caption{Post-hoc failure attribution taxonomy and log-evidence criteria supporting the failure analysis in \cref{sec:failures}.}
\label{tab:failure_taxonomy}
\begin{tabularx}{\textwidth}{@{}L{0.19\textwidth}YY@{}}
\toprule
Category & Log / trace evidence & Exclusion \\
\midrule
\multicolumn{3}{@{}l}{\textit{Single-robot}} \\
Visual grounding / search & Wrong source, object, or destination, or an unsupported progress claim & A correct target that is physically unreachable is not a perception error \\
Navigation & Failure to reach an operating stance before an interaction & Not contention unless teammate-related evidence exists \\
Acquisition & An attempted grasp or pickup fails its physical criterion (\eg, IK or arm-path rejection) & An unreached stage is not an attempted failure \\
Placement / release & An attempted placement fails support, containment, or stability & Later loss caused by the teammate belongs to a coordination category \\
\multicolumn{3}{@{}l}{\textit{Coordination}} \\
Workspace contention & The teammate's occupancy obstructs an intended action, with a time-aligned trace & Proximity or waiting alone is not contention \\
Dependency timing & The consumer acts before the required condition holds & A different valid order is not a timing error \\
Unusable produced state & The producer's action completes, but the condition required by the consumer is absent, \eg, an unstable or unreachable release & A single failed receiver does not prove general inaccessibility \\
Payload loss in transport & Loss of support or containment during carrier transport & Empty-carrier navigation cannot diagnose this category \\
Stale shared state & The team acts on an invalidated belief or message, or fails to recover after contrary evidence & Hidden beliefs are not inferred without messages or decisions \\
\midrule
Infrastructure incident & Host, service, or simulator failure logged under the incident policy (Sec.~D.2) & Reruns logged separately; never counted as policy failure \\
\bottomrule
\end{tabularx}
\end{table}

\section{Release and Provenance}
\label{supp:release}

\subsection{Released artifacts}
The release includes task packages and splits with seed configurations, the qualification records and reference-trace hashes, evaluator definitions and regression tests, public observation and action schemas, the adapted baseline revisions, tool and checkpoint configurations, development-artifact hashes, and rollout event logs. Licensed simulator and scene assets are distributed only as permitted, with documented acquisition and version checks.

Every reported result is bound to a manifest containing the task set, robots, sensor profile, localization source, declared assistance, physics settings, role information, development and test budgets, reset rules, and scoring version. A changed manifest produces a different configuration name. Scores on the test split are reported together with its selection rule, family coverage, and embodiment pairings.

\subsection{Component provenance}
The early $\gamma$-Scape implementation supplies the graspable small-object assets, the BenchForge task-package format, and the base single-robot Oracle architecture~\citep{zhang2026gammascape}. VLNVerse supplies the scenes adapted for this benchmark~\citep{vlnverse}. \bench adds the TidyBot--ARX5 and Create~3 configurations and their Oracle adapters, the team-conditioned generators for homogeneous and heterogeneous teams, joint qualification, the evaluator for cross-robot dependencies, and the two-level evaluation protocol.

\subsection{Criteria and evidence for the benchmark comparison}
\label{supp:comparison}
Each cell of \cref{tab:comparison} is judged on the main evaluation setting described in the cited paper; optional variants, such as oracle-perception or non-privileged-perception ablations, do not change a cell. The columns are binary. \emph{MR}: two or more embodied agents act in one scene. \emph{MM}: the evaluated agents are responsible for both navigation and object pick-and-place. \emph{Het.}: the agents have different embodiments. \emph{Phys.\ dep.}: a task requires one agent to act on an object state produced by another agent, that state results from simulated physics rather than from a scripted, snapped, or symbolic state change, and success is checked from object state. \emph{Sensor}: policies observe only sensor data and language, without simulator-provided object identities or poses.
\begin{itemize}[leftmargin=*,itemsep=1pt]
    \item \textbf{FurnMove}~\citep{cordialsync}: two identical agents start holding a lifted object and move it with discrete actions; observations are egocentric RGB images. There is no pick-and-place and no producer--consumer dependency.
    \item \textbf{TDW-MAT}~\citep{coela}: two agents of the same embodiment transport objects with containers using navigation and grasp actions; observations are egocentric RGB-D images. No task requires one agent to use a state produced by the other.
    \item \textbf{Habitat 3.0}~\citep{habitat3}: a Spot robot and a humanoid rearrange objects. Object positions and goals are given as 3D coordinates, and the robot observes the humanoid's relative distance and heading. Tasks impose no required cross-agent dependency.
    \item \textbf{PARTNR}~\citep{partnr}: a Spot robot and a humanoid act through high-level planners and skills; the main setting uses a world graph built from ground-truth annotations. Constraints concern task order and agent capability, and its oracle skills use privileged simulation information.
    \item \textbf{EMOS (Habitat-MAS)}~\citep{emos}: Fetch, Stretch, Spot, and drone robots execute skills implemented with ground-truth world information; objects snap to the gripper with contact physics disabled, and agents receive a textual scene description.
    \item \textbf{CoEmbod}~\citep{coembod}: two executors with identical action spaces act in OmniGibson through discrete high-level primitives; dependencies such as synchronized lifting, unlocking, and restarting are realized through these primitives, and observations include a simulator-provided list of visible objects.
    \item \textbf{CoCoBench}~\citep{cocobench}: homogeneous agents in AI2-THOR act through parameterized high-level skills; handoffs pass objects through intermediate receptacles and are scored from skill execution events, and action menus are built from simulator metadata.
    \item \textbf{RoCoBench}~\citep{roco}: UR5e, Franka Panda, and humanoid arms share a tabletop in MuJoCo. In Make Sandwich and Sort Cubes, robots stack on or move objects for a teammate, and success is checked from object states; agents receive object locations rather than images.
    \item \textbf{RoboFactory}~\citep{robofactory}: identical Franka Panda arms in ManiSkill, as specified in the released task configurations; tasks such as passing a shoe require object transfer between arms, and policies act on camera images.
    \item \textbf{BEHAVIOR-1K}~\citep{behavior}: single-robot household activities in OmniGibson with image observations.
    \item \textbf{CoNavBench}~\citep{conavbench}: two robots perform relay-style collaborative navigation in Habitat with RGB-D observations and the atomic actions move forward, turn, and stop. Object pickup, handoff, and delivery are specified at the task-planning level but not executed as manipulation, and results are reported separately for Fetch and Spot configurations.
    \item \textbf{DeCoNav}~\citep{deconav}: Fetch and Spot robots perform collaborative navigation without manipulation, using RGB-D observations.
\end{itemize}
ManiSkill-HAB~\citep{maniskillhab} is omitted from the table because it evaluates a single robot, replaces navigation with teleporting, and includes ground-truth target poses in its observations.

\section{Additional Results}
\label{supp:results}

\begin{table}[!b]
\centering\small
\caption{Full end-to-end results by team type (\%) and cost (model calls per episode, summed over all team members: two robots in 98 tasks, three in the two two-ended tasks). All rows use GPT-5.6-luna except the last; +prior and \method use the scene prior, and \method also its navigation memory and skill library. HSR is given as successful over attempted handoffs; --: no handoff attempted. \Cref{tab:main} reports these metrics over all tasks.}
\label{tab:e2e_full}
\setlength{\tabcolsep}{3.2pt}
\begin{tabular*}{\textwidth}{@{\extracolsep{\fill}}l cccccc cccccc c@{}}
\toprule
 & \multicolumn{6}{c}{Homogeneous (75 tasks)} & \multicolumn{6}{c}{Heterogeneous (25 tasks)} & Cost \\
\cmidrule(lr){2-7}\cmidrule(lr){8-13}
Method & SR & GC & NSR & PE & HCov & HSR & SR & GC & NSR & PE & HCov & HSR & calls/ep. \\
\midrule
ASPIRE-SG & \res{0.0} & \res{4.2} & \res{16.3} & \res{14.7} & \res{1.2} & -- & \res{0.0} & \res{0.0} & \res{0.0} & \res{0.0} & \res{0.0} & -- & \res{20.4} \\
RAX-SG & \res{0.0} & \res{3.8} & \res{15.4} & \res{13.2} & \res{0.0} & -- & \res{0.0} & \res{0.0} & \res{0.0} & \res{0.0} & \res{0.0} & -- & \res{21.3} \\
ASPIRE-SG + prior & \res{0.0} & \res{10.1} & \res{51.0} & \res{42.3} & \res{2.5} & -- & \res{0.0} & \res{8.0} & \res{23.5} & \res{22.3} & \res{0.0} & -- & \res{20.4} \\
RAX-SG + prior & \res{0.0} & \res{11.8} & \res{53.8} & \res{44.4} & \res{6.2} & \res{0/1} & \res{0.0} & \res{4.3} & \res{11.8} & \res{11.1} & \res{0.0} & -- & \res{20.0} \\
\method & \res{0.0} & \res{13.5} & \res{59.6} & \res{52.9} & \res{12.3} & \res{1/5} & \res{0.0} & \res{12.3} & \res{43.1} & \res{40.3} & \res{0.0} & -- & \res{31.2} \\
\midrule
\method (GPT-6.1-sol) & \res{0.0} & \res{20.5} & \res{67.3} & \res{57.5} & \res{19.8} & \res{3/5} & \res{4.0} & \res{24.6} & \res{47.1} & \res{40.2} & \res{33.3} & \res{1/6} & \res{24.7} \\
\bottomrule
\end{tabular*}
\end{table}

\paragraph{Full results.}
\Cref{tab:e2e_full} breaks the end-to-end results down by team type and adds cost; the diagnostic by stage and condition is reported in \cref{tab:diag}.

\clearpage
\begin{table}[tb]
\centering\small
\caption{End-to-end SR / GC (\%) of \method by task family, with GPT-5.6-luna and GPT-6.1-sol.}
\label{tab:per_family}
\setlength{\tabcolsep}{2.7pt}
\begin{tabular}{@{}lccc@{}}
\toprule
Family & Tasks & GPT-5.6-luna & GPT-6.1-sol \\
\midrule
Multi-destination distribution & 8 & \res{0.0} / \res{20.3} & \res{0.0} / \res{32.8} \\
Multi-source collection & 7 & \res{0.0} / \res{7.1} & \res{0.0} / \res{8.9} \\
Container packing & 12 & \res{0.0} / \res{11.1} & \res{0.0} / \res{17.4} \\
Object exchange & 33 & \res{0.0} / \res{11.7} & \res{0.0} / \res{17.0} \\
Clear route and deliver & 8 & \res{0.0} / \res{21.1} & \res{0.0} / \res{26.9} \\
Environment restoration & 7 & \res{0.0} / \res{16.5} & \res{0.0} / \res{32.3} \\
\midrule
Two-source collection & 4 & \res{0.0} / \res{6.7} & \res{0.0} / \res{11.8} \\
Clear then deliver & 2 & \res{0.0} / \res{7.1} & \res{0.0} / \res{17.9} \\
High-to-low relay & 4 & \res{0.0} / \res{0.0} & \res{0.0} / \res{0.0} \\
Low-to-high relay & 3 & \res{0.0} / \res{0.0} & \res{0.0} / \res{0.0} \\
Narrow-passage relay & 2 & \res{0.0} / \res{0.0} & \res{0.0} / \res{44.0} \\
Multi-object carrier transport & 8 & \res{0.0} / \res{30.8} & \res{12.5} / \res{50.4} \\
Two-ended loading/unloading & 2 & \res{0.0} / \res{10.0} & \res{0.0} / \res{20.0} \\
\bottomrule
\end{tabular}
\end{table}

\begin{table}[!b]
\centering\small
\caption{Handoff-isolated diagnostic by dependency type: number of full-condition trials in which the producer establishes the condition ($E_d$) / the handoff is completed (DSR). Relays include the surface handoffs of object exchange; settings (a)--(d) follow \cref{tab:diag}.}
\label{tab:diag_type}
\setlength{\tabcolsep}{3pt}
\begin{tabular}{@{}llcccc@{}}
\toprule
 & Method & Relay & Carrier & Clearance & Unloading \\
\midrule
\multirow{3}{*}{(a)} & ASPIRE-SG & \res{2} / \res{0} & \res{0} / \res{0} & \res{2} / \res{0} & \res{0} / \res{0} \\
 & RAX-SG & \res{2} / \res{0} & \res{0} / \res{0} & \res{0} / \res{0} & \res{0} / \res{0} \\
 & \method & \res{0} / \res{0} & \res{0} / \res{0} & \res{0} / \res{0} & \res{0} / \res{0} \\
\midrule
\multirow{3}{*}{(b)} & ASPIRE-SG & \res{0} / \res{0} & \res{0} / \res{0} & \res{0} / \res{0} & \res{0} / \res{0} \\
 & RAX-SG & \res{0} / \res{0} & \res{0} / \res{0} & \res{1} / \res{0} & \res{0} / \res{0} \\
 & \method & \res{1} / \res{0} & \res{0} / \res{0} & \res{3} / \res{0} & \res{0} / \res{0} \\
\midrule
\multirow{3}{*}{(c)} & ASPIRE-SG & \res{0} / \res{0} & \res{2} / \res{0} & \res{2} / \res{0} & \res{0} / \res{0} \\
 & RAX-SG & \res{5} / \res{0} & \res{1} / \res{0} & \res{1} / \res{0} & \res{0} / \res{0} \\
 & \method & \res{4} / \res{1} & \res{10} / \res{0} & \res{12} / \res{3} & \res{0} / \res{0} \\
\midrule
\multirow{3}{*}{(d)} & ASPIRE-SG & \res{11} / \res{5} & \res{7} / \res{2} & \res{15} / \res{6} & \res{0} / \res{0} \\
 & RAX-SG & \res{16} / \res{7} & \res{7} / \res{2} & \res{15} / \res{5} & \res{1} / \res{0} \\
 & \method & \res{16} / \res{11} & \res{10} / \res{7} & \res{28} / \res{15} & \res{1} / \res{0} \\
\midrule
 & Trials & \res{35} & \res{18} & \res{43} & \res{2} \\
\bottomrule
\end{tabular}
\end{table}

\paragraph{Families and dependency types.}
\Cref{tab:per_family} breaks down end-to-end performance across all 13 task families. With GPT-5.6-luna, no method credits a checkpoint on the three relay families, whereas multi-object carrier transport is the most tractable heterogeneous family and contains the only successful episode, with GPT-6.1-sol. \Cref{tab:diag_type} further breaks down the 98-trial handoff-isolated diagnostic by dependency type. In the unguided full diagnostic (c) with GPT-6.1-sol, \method establishes the condition in 4/35 relay, 10/18 carrier-loading, 12/43 clearance, and 0/2 unloading trials; none of the 18 carrier-loading trials in (c) reaches diagnostic completion, whereas 7/18 succeed once low-level inputs are equalized in (d). With Oracle guidance (d), \method completes more relay (11 \vs at most 7), carrier (7 \vs 2), and clearance (15 \vs at most 6) handoffs than both baselines.

\clearpage
\begin{table}[tb]
\centering\small
\caption{End-to-end results (\%) by coordination mechanism and embodiment. P: parallel only ($\mathcal D=\varnothing$); D: with physical dependency. All rows use GPT-5.6-luna except those marked GPT-6.1-sol. HSR is given as successful over attempted handoffs; --: no dependency or no handoff attempted.}
\label{tab:mechanism}
\setlength{\tabcolsep}{4pt}
\begin{tabular*}{\textwidth}{@{\extracolsep{\fill}}lllccccc@{}}
\toprule
Embodiment & Mechanism & Method & SR & GC & NSR & HCov & HSR \\
\midrule
\multirow{12}{*}{Homogeneous} & \multirow{6}{*}{P (34 tasks)} & ASPIRE-SG & \res{0.0} & \res{6.2} & \res{18.5} & -- & -- \\
 &  & RAX-SG & \res{0.0} & \res{4.7} & \res{14.8} & -- & -- \\
 &  & ASPIRE-SG + prior & \res{0.0} & \res{13.2} & \res{51.9} & -- & -- \\
 &  & RAX-SG + prior & \res{0.0} & \res{16.2} & \res{63.0} & -- & -- \\
 &  & \method & \res{0.0} & \res{13.6} & \res{55.6} & -- & -- \\
 &  & \method (GPT-6.1-sol) & \res{0.0} & \res{22.3} & \res{59.3} & -- & -- \\
\cmidrule(l){2-8}
 & \multirow{6}{*}{D (41 tasks)} & ASPIRE-SG & \res{0.0} & \res{2.6} & \res{14.0} & \res{1.2} & -- \\
 &  & RAX-SG & \res{0.0} & \res{3.1} & \res{16.0} & \res{0.0} & -- \\
 &  & ASPIRE-SG + prior & \res{0.0} & \res{7.6} & \res{50.0} & \res{2.5} & -- \\
 &  & RAX-SG + prior & \res{0.0} & \res{8.1} & \res{44.0} & \res{6.2} & \res{0/1} \\
 &  & \method & \res{0.0} & \res{13.5} & \res{64.0} & \res{12.3} & \res{1/5} \\
 &  & \method (GPT-6.1-sol) & \res{0.0} & \res{18.9} & \res{76.0} & \res{19.8} & \res{3/5} \\
\midrule
\multirow{12}{*}{Heterogeneous} & \multirow{6}{*}{P (4 tasks)} & ASPIRE-SG & \res{0.0} & \res{0.0} & \res{0.0} & -- & -- \\
 &  & RAX-SG & \res{0.0} & \res{0.0} & \res{0.0} & -- & -- \\
 &  & ASPIRE-SG + prior & \res{0.0} & \res{0.0} & \res{0.0} & -- & -- \\
 &  & RAX-SG + prior & \res{0.0} & \res{0.0} & \res{0.0} & -- & -- \\
 &  & \method & \res{0.0} & \res{6.7} & \res{33.3} & -- & -- \\
 &  & \method (GPT-6.1-sol) & \res{0.0} & \res{11.8} & \res{33.3} & -- & -- \\
\cmidrule(l){2-8}
 & \multirow{6}{*}{D (21 tasks)} & ASPIRE-SG & \res{0.0} & \res{0.0} & \res{0.0} & \res{0.0} & -- \\
 &  & RAX-SG & \res{0.0} & \res{0.0} & \res{0.0} & \res{0.0} & -- \\
 &  & ASPIRE-SG + prior & \res{0.0} & \res{9.5} & \res{30.8} & \res{0.0} & -- \\
 &  & RAX-SG + prior & \res{0.0} & \res{5.1} & \res{15.4} & \res{0.0} & -- \\
 &  & \method & \res{0.0} & \res{13.4} & \res{46.2} & \res{0.0} & -- \\
 &  & \method (GPT-6.1-sol) & \res{4.8} & \res{27.0} & \res{51.3} & \res{33.3} & \res{1/6} \\
\bottomrule
\end{tabular*}
\end{table}

\paragraph{Mechanism and embodiment.}
\Cref{tab:mechanism} breaks down the end-to-end results in \cref{tab:main} by mechanism and embodiment. At matched embodiment, physical dependencies roughly halve the GC of the +prior baselines on homogeneous tasks (\eg, from \res{16.2}\% to \res{8.1}\% for RAX-SG + prior), whereas \method's GC remains stable (\res{13.6}\% \vs \res{13.5}\%). On heterogeneous tasks, \method achieves the highest GC and NSR on both the parallel-only (4 tasks) and dependency (21 tasks) subsets.

\end{document}